\documentclass[]{bytedance}
\usepackage[toc,page,header]{appendix}

\usepackage{minitoc}
\usepackage{amsfonts}
\usepackage{amssymb}
\usepackage{tabularx}
\usepackage{listings}
\usepackage{xcolor}
\usepackage{cancel}

\usepackage{tabulary,multirow,xspace}
\usepackage{fixmath,mathtools,nicefrac,mmstyle}
\usepackage{subcaption}
\usepackage{caption}
\usepackage{wrapfig} 
\usepackage[misc]{ifsym} 
\usepackage{colortbl}

\usepackage{wrapfig}
\usepackage{multicol}
\usepackage[most]{tcolorbox}
\usepackage{pifont}

\definecolor{codegreen}{rgb}{0,0.6,0}
\definecolor{codegray}{rgb}{0.5,0.5,0.5}
\definecolor{codepurple}{rgb}{0.58,0,0.82}
\definecolor{backcolour}{rgb}{0.95,0.95,0.92}
\definecolor{boxblue}{RGB}{57,89,163}
\definecolor{boxbluebg}{RGB}{230,237,250} 

\lstdefinestyle{mystyle}{
    backgroundcolor=\color{backcolour},   
    commentstyle=\color{codegreen},
    keywordstyle=\color{magenta},
    numberstyle=\tiny\color{codegray},
    stringstyle=\color{codepurple},
    basicstyle=\ttfamily\footnotesize,
    breakatwhitespace=false,         
    breaklines=true,                 
    captionpos=b,                    
    keepspaces=true,                 
    numbers=none,                    
    numbersep=5pt,                  
    showspaces=false,                
    showstringspaces=false,
    showtabs=false,                  
    tabsize=2
}
\definecolor{mygray1}{gray}{.95}
\definecolor{mygray2}{gray}{.9}
\definecolor{mygray3}{gray}{.95}
\usepackage{pifont}

\newlength\savewidth
\newcolumntype{x}[1]{>{\centering\arraybackslash}p{#1pt}}

\newcommand{\app}{\raise.17ex\hbox{$\scriptstyle\sim$}}

\usepackage{xcolor}
\usepackage{graphicx}
\usepackage{amssymb}
\usepackage{pifont}
\usepackage{floatrow}
\usepackage{amsmath} 
\usepackage{float}
\usepackage{wrapfig}
\usepackage{multirow}
\usepackage{tcolorbox}
\tcbuselibrary{breakable, skins, raster}
\usepackage{listings}
\usepackage{listings}

\definecolor{carnelian}{rgb}{0.7, 0.11, 0.11}
\usepackage{booktabs}
\usepackage{makecell}
\usepackage{multirow}

\usepackage{adjustbox}
\usepackage{cuted} 

\usepackage{algorithm}
\usepackage{algorithmic}
\definecolor{myblue}{RGB}{210, 225, 255}
\definecolor{mytextblue}{RGB}{51, 161, 201}
\definecolor{mypurple}{RGB}{218, 112, 214}

\definecolor{commentgreen}{rgb}{0.1, 0.4, 0.1}
\definecolor{keywordblue}{rgb}{0.1, 0.1, 0.7}
\definecolor{stringred}{rgb}{0.7, 0.1, 0.1}

\lstdefinestyle{mystyle}{
    commentstyle=\color{commentgreen},
    keywordstyle=\color{keywordblue},   
    stringstyle=\color{stringred},
    basicstyle=\ttfamily\scriptsize, 
    breaklines=true,
    keepspaces=true,
    showstringspaces=false,
    frame=none,                     
    language=Python, 
}

\title{OmniTransfer: All-in-one Framework for Spatio-temporal Video Transfer}

\author{
\centerline{
Pengze Zhang \quad
Yanze Wu$^{\dagger}$ \quad 
Mengtian Li \quad
Xu Bai \quad
Songtao Zhao$^{\ddagger}$ \quad
}
\centerline{
Fulong Ye \quad
Chong Mou \quad
Xinghui Li \quad
Zhuowei Chen \quad
Qian He \quad
Mingyuan Gao
}
}

\affiliation[]{Intelligent Creation Lab, ByteDance}

\abstract{
Videos convey richer information than images or text, capturing both spatial and temporal dynamics. However, most existing video customization methods rely on reference images or task-specific temporal priors, failing to fully exploit the rich spatio-temporal information inherent in videos, thereby limiting flexibility and generalization in video generation. To address these limitations, we propose OmniTransfer, a unified framework for spatio-temporal video transfer. It leverages multi-view information across frames to enhance appearance consistency and exploits temporal cues to enable fine-grained temporal control. To unify various video transfer tasks, OmniTransfer incorporates three key designs: Task-aware Positional Bias that adaptively leverages reference video information to improve temporal alignment or appearance consistency; Reference-decoupled Causal Learning separating reference and target branches to enable precise reference transfer while improving efficiency; and Task-adaptive Multimodal Alignment using multimodal semantic guidance to dynamically distinguish and tackle different tasks. Extensive experiments show that OmniTransfer outperforms existing methods in appearance (ID and style) and temporal transfer (camera movement and video effects), while matching pose-guided methods in motion transfer without using pose, establishing a new paradigm for flexible, high-fidelity video generation.
}

\date{January 20, 2026}
\checkdata[Project Page]{\url{https://pangzecheung.github.io/OmniTransfer/}}

\begin{document}
\maketitle

\renewcommand{\thefootnote}{\fnsymbol{footnote}}
\footnotetext{\parbox[t]{\linewidth}{$\dagger$ Corresponding author, $\ddagger$ Project lead. 
This work is purely academic and non-commercial. Demo reference images/videos\\
are from public domains or AI-generated. For copyright concerns, please contact us for the removal of relevant content.}}
\renewcommand{\thefootnote}{\arabic{footnote}}

\section{Introduction}
\label{sec:intro}

We have all heard the old adage: \textit{``A picture is worth a thousand words''}—and if we follow that logic a little further, a video might be worth a million. After all, a static image freezes just one moment of light, texture, and form; video weaves those moments into something dynamic, carrying not just how things look, but how they move and change. These are details that neither words nor a single image can fully convey. 

The same insight applies to diffusion-based video generation. A reference video offers greater flexibility and capability than using an image or text alone. Not only can the model reference identity (ID) and style in spatial appearance aspects, but it can also exploit temporal information, including camera movement, motion, and visual effects, enabling more coherent and expressive video synthesis.

\begin{figure*}[!t]
    \centering
    \includegraphics[width=0.97\textwidth]{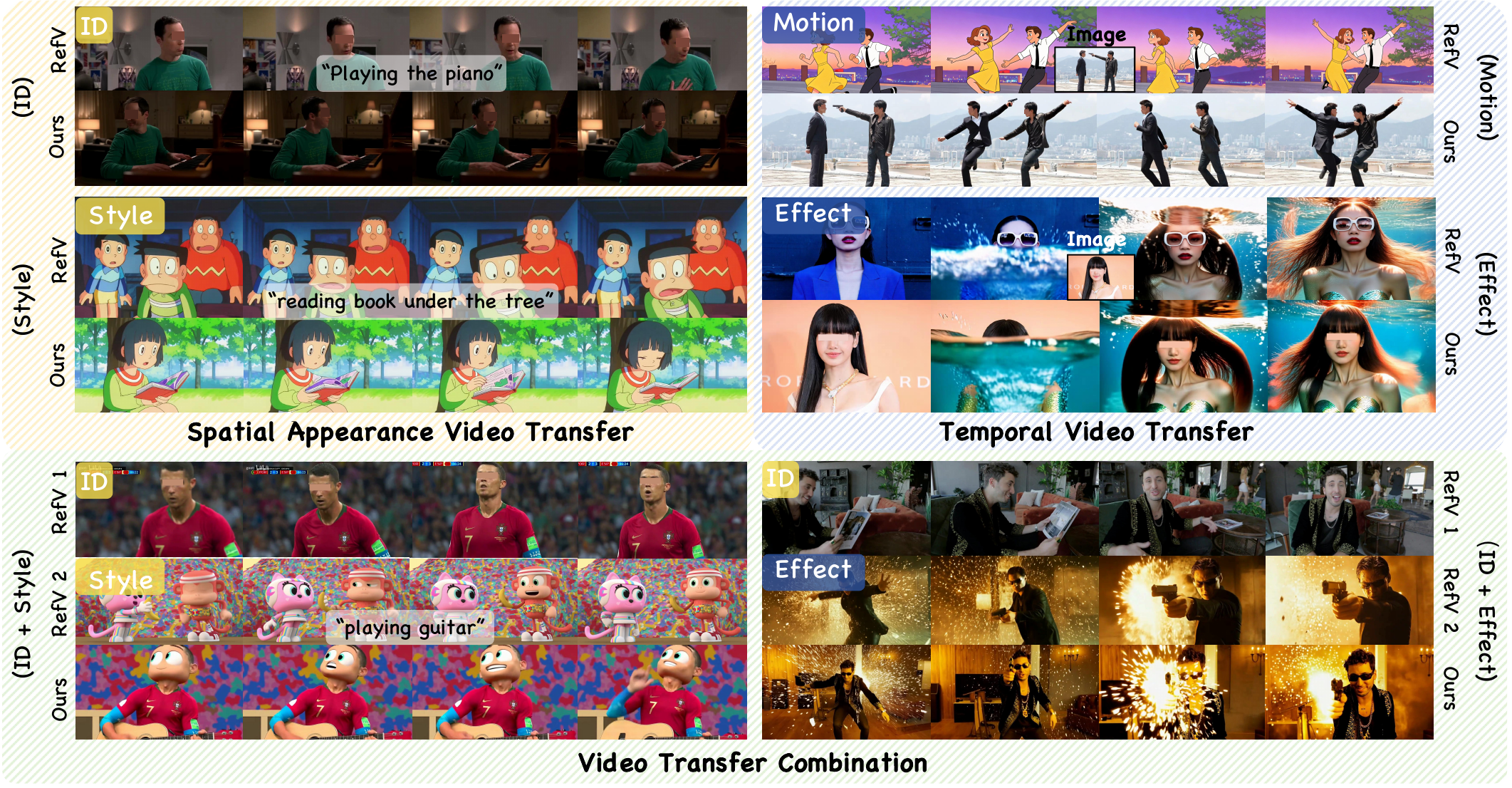}
    \caption{ OmniTransfer seamlessly unifies spatial appearance (ID and style) and temporal video transfer tasks (effect, motion and camera movement) within a single framework, and exhibits strong generalization across unseen task combinations.}
    \label{fig:abstract}
\end{figure*} 


Despite this potential, existing methods have yet to fully exploit their spatio-temporal information. 1) For spatial information, most approaches~\cite{consistid,liu2025phantom,standin,stylemaster} still mainly rely on pixel-level cues extracted from reference images. However, cues such as ID and style are inherently spatio-temporal, extending across multiple frames and views. This makes it difficult for a single image to capture their full details, thereby limiting their effectiveness. 2) For temporal information, current approaches remain in its early stages: some depend on pre-defined physical priors (e.g., pose or camera position)~\cite{zheng2024cami2v,hou2025trainingfreecameracontrolvideo,wananimate,mimicmotion}, some employ inversion-based reconstruction~\cite{hu2024motionmaster,motionclone}, while others require additional fine-tuning for specific temporal dynamics during test time~\cite{10.1007/978-3-031-72992-8_16,pondaven2025videomotiontransferdiffusion}. Recently, CamCloneMaster~\cite{camclonemaster} made an initial reference-based attempt on camera motion through temporal context concatenation, yet it struggles to generalize to in-the-wild conditions, and fails to address general temporal video reference tasks.
These limitations highlight the need for a unified framework that directly leverages temporal information from reference videos, enabling flexible and robust video customization in real-world scenarios.

To tackle the aforementioned challenges, this paper introduces a novel exploration in the domain of video reference, proposing an all-in-one framework for spatio-temporal video transfer, OmniTransfer. This framework not only integrates multi-frame reference information from the reference video at the spatial level, enhancing the consistency of reference video ID and style, but also effectively models temporal aspects such as motion, camera movement, and video effects, providing multi-dimensional control and unification over the video generation process. 
First, to unify video transfer tasks, we introduce Task-aware Positional Bias. 
For temporal transfer tasks, based on our assumption that video diffusion models inherently maintain temporal consistency through spatial context, we add spatial offsets to positional embeddings to preserve temporal alignment.
For appearance transfer tasks, temporal offsets are applied to propagate appearance information across frames.
Second, we introduce Reference-decoupled Causal Learning, which employs unidirectional transfer from reference to target, preventing simple copy-pasting. By separating the reference and target branches, the reference branch requires only a single forward pass, reducing computational time by 20\% compared to full-attention models.
Finally, to improve semantic guidance and avoid cross-task confusion, we introduce a Multimodal Large Language Model (MLLM) via a Task-adaptive Multimodal Alignment module. This module leverages multiple sets of task-specific MetaQueries~\cite{pan2025transfer} to dynamically integrate and align semantic information across tasks, effectively meeting the requirements of each task.

In sum, our contribution can be summarized as follows.
\begin{itemize}
    \item We propose OmniTransfer, a unified framework for the new task of spatio-temporal video transfer that supports diverse tasks such as identity, style, effect, camera movement, and motion transfer, while generalizing seamlessly to their compositional combinations (Fig.~\ref{fig:abstract}).
    \item We propose Task-aware Positional Bias, Reference-decoupled Causal Learning, and Task-adaptive Multimodal Alignment, each designed to unify various video customization tasks, enabling efficient and flexible spatio-temporal information transfer.
    \item Experiments show that OmniTransfer outperforms existing methods in appearance (ID, style) and temporal (camera movement, effects) transfer, while matching pose-guided methods in motion transfer without using pose.
    Moreover, benefiting from our model design, it achieves these improvements with a 20\% reduction in runtime compared to the basemodel architecture.
\end{itemize}

\section{Related Work}
\subsection{Appearance reference task}

The two main tasks in appearance reference are ID and style transfer.
For ID transfer in images, approaches have evolved from adapter-based tuning~\cite{ye2023ip,guo2024pulid,wang2024instantid,jiang2025infiniteyou,dreamid,Li_2025_CVPR} to in-context learning~\cite{dreamo,xiao2025omnigen,chen2025xverse,musar,anydressing}.
In video ID transfer, most approaches still rely on single reference images for ID preservation~\cite{ID-Animator,consistid,standin,dreamidv}, while some works further explore broader subject-to-video generation tasks~\cite{hu2025hunyuancustom,huang2025conceptmaster,liang2025movie,chen2025multi,liu2025phantom,omniinsert,instructx}.
For instance, Phantom~\cite{liu2025phantom} concatenates ID features along the temporal dimension to maintain appearance consistency.
Similarly, style transfer has been widely explored in image diffusion models~\cite{instantstyle,instantstyleplus,styletokenizer,csgo}.
Early video stylization extends these models to the temporal domain via Image-to-Video paradigms~\cite{anyv2v,geyer2023tokenflow,stillmoving}, while recent Text-to-Video methods~\cite{videocrafter1,wang2024videocomposer,stylecrafter,stylemaster} offer more flexible and controllable stylization.
Despite these advances, these methods rely solely on single-image references, overlooking the rich multi-frame and multi-view cues inherent in videos. Our approach exploits these cues to achieve consistent appearance and stable temporal coherence.

\begin{figure*}[!t]
    \centering
    \includegraphics[width=\textwidth]{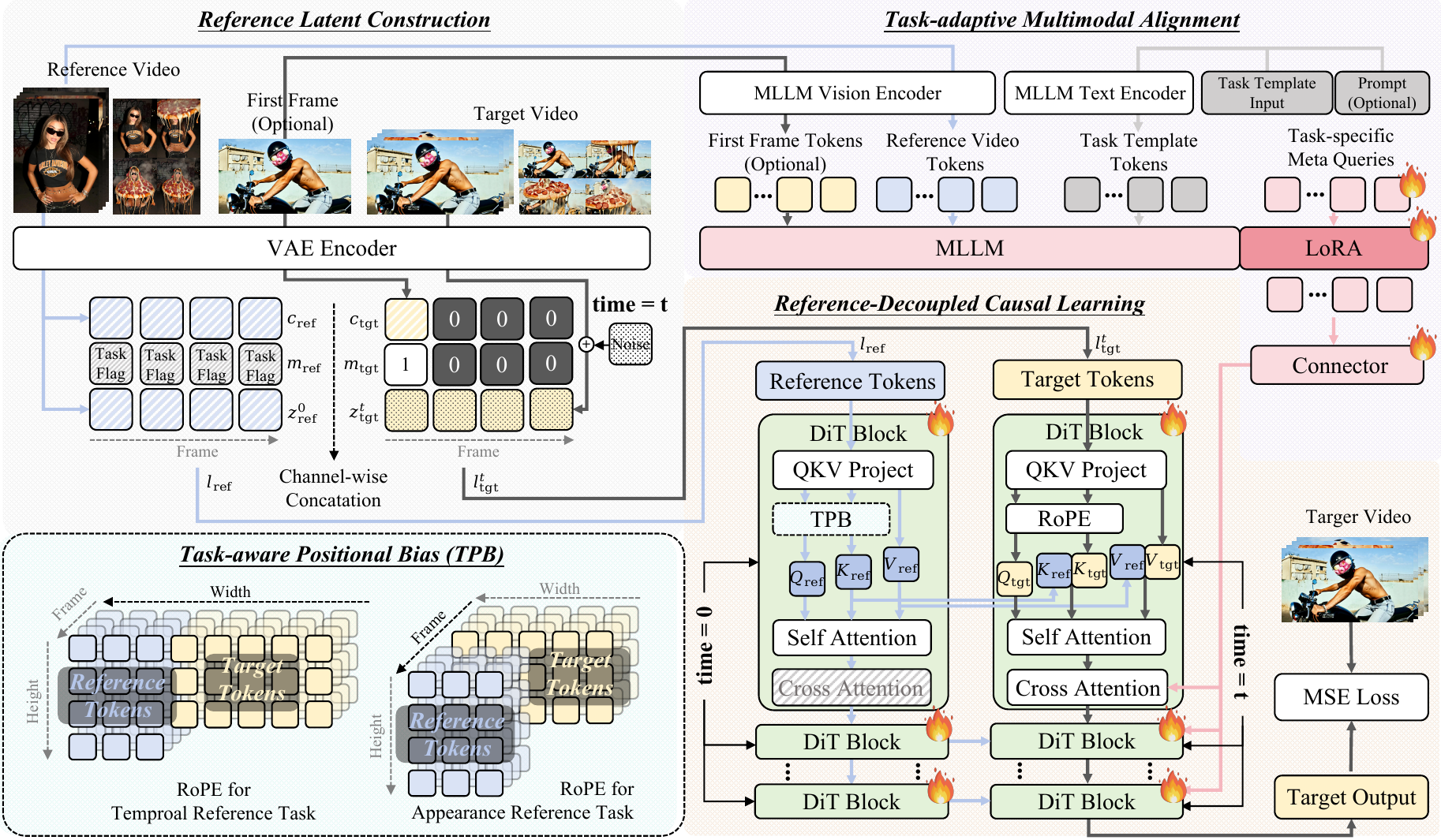}
    \caption{OmniTransfer comprises three key components: 1) Task-aware Positional Bias: exploits the model’s inherent spatial and temporal context capabilities for diverse tasks. 2) Reference-decoupled Causal Learning: separates reference and target branches for causal and efficient transfer. 3) Task-adaptive Multimodal Alignment: leverages an MLLM to unify and enhance semantic understanding across tasks.}
    \label{fig:omnitransfer}
\end{figure*}

\subsection{Temporal reference task}
Temporal video customization covers motion generation, camera movement, and effect synthesis.
Pose motion transfer initially relies on GAN-based warping~\cite{fomm,tpsmm,mraa,Zhang_2022_CVPR}, while diffusion-based models~\cite{aa,magicanimate,magicpose,champ,stableanimator,unianimatedit,realisdancedit,dreamactorm1,mimicmotion,wananimate} improve temporal smoothness.
However, they are limited by pose priors, as skeleton alignment may cause loss of appearance cues and hinder their extension to multi-person scenarios.
Some studies explore more general motion control~\cite{10.1007/978-3-031-72992-8_16,Jeong_2024_CVPR,Yatim_2024_CVPR,10.1145/3721238.3730683,10.1145/3721238.3730735,ma2025follow,pondaven2025videomotiontransferdiffusion}, but most rely on diffusion inversion or test-time finetuning.
Camera motion generation typically uses explicit parameterization~\cite{wang2024motionctrl,xu2024camco,he2025cameractrl,li2025realcami2v,zheng2024cami2v,hou2025trainingfreecameracontrolvideo,wang2025cinemaster} and parameter-free attention inversion~\cite{hu2024motionmaster,motionclone}, while CamCloneMaster~\cite{camclonemaster} explores temporal context concatenation, yet resolution and generalization remain limited.
Effect generation in industry often relies on LoRA-based~\cite{hu2022lora} finetuning, whereas academic works~\cite{magicvfx,omnieffects,vfxcreater} focus on precise spatial and temporal control of effects, but cannot generate new effects conditioned on a reference video.
In contrast, our OmniTransfer achieves unified motion, camera, and effect transfer without explicit priors, demonstrating strong generalization to in-the-wild scenarios.

\section{Preliminary}
\label{sec:preliminary}
Our framework is built upon Wan2.1 I2V 14B~\cite{wan2025} as the underlying diffusion model.
The input latent $l^t\in \mathbb{R}^{f\times h \times w \times (2n+4)} = [c,m,z^t]$ concatenates three components along the channel dimension: latent noise $z^t \in \mathbb{R}^{f\times h \times w \times n}$ obtained by adding timestep $t$ noise to VAE-compressed video features $z$; condition latent $c \in \mathbb{R}^{f\times h \times w \times n}$ encoded from the condition image $I$ concatenated with zero-filled frames; binary mask latent $m \in \mathbb{R}^{f\times h \times w \times 4}$ with values of 1 for preserved and 0 for generated frames. ${[\cdot,\cdot]}$ denotes feature concatenation along the channel dimension, and $f$, $h$, $w$, $n$ represent frame number, height, width, and channel dimension, respectively.

Each Diffusion Transformer (DiT) block in Wan2.1 includes self-attention and cross-attention layers. The self-attention adopts 3D Rotary Positional Embedding (RoPE):
\begin{equation}
\text{Attn}(R_\theta(Q), R_\theta(K), V) = 
\text{softmax}\!\left(
    \frac{R_\theta(Q)\, R_\theta(K)^{\top}}
         {\sqrt{d}}
\right)V,
\end{equation}
where \(Q = W_Q l^t\), \(K = W_K l^t\), \(V = W_V l^t\). \(R_\theta(\cdot)\) denotes the RoPE rotation applied to queries and keys, $W_Q$, $W_K$, $W_V$ are learnable projections, and \(d\) is the feature dimension.
Cross-attention integrates textual features as
$\text{Attn}(Q,K_\text{p},V_\text{p})$ with \(K_\text{p}\) and \(V_\text{p}\) derived from prompt $p$.


\section{Method}
\setlength{\tabcolsep}{28pt} 
\begin{table}[t]
\centering
\renewcommand{\arraystretch}{1}
\begin{tabular}{c c c}
\toprule
\textbf{Task} & \textbf{Input} & \textbf{Output}-$V_{tgt}$ \\
\midrule
\multicolumn{3}{c}{\textbf{Appearance Transfer (T2V)}} \\
\midrule
ID Transfer & $V_\text{ref}, p$ & ID from $V_\text{ref}$, following prompt $p$ \\
Style Transfer & $V_\text{ref}, p$ & Style from $V_\text{ref}$, following prompt $p$ \\
\midrule
\multicolumn{3}{c}{\textbf{Temporal Transfer (I2V)}} \\
\midrule
Motion Transfer & $V_\text{ref}, I$ & Motion from $V_\text{ref}$, starting from $I$ \\
Camera Movement & $V_\text{ref}, I$ & Camera movement from $V_\text{ref}$, starting from $I$ \\
Effect Transfer & $V_\text{ref}, I$ & Effect from $V_\text{ref}$, starting from $I$ \\
\bottomrule
\end{tabular}
\caption{
Overview of video reference tasks. $V_\text{ref}$: reference video; $I$: first-frame image; $p$: prompt; $V_\text{tgt}$: generated video. 
}
\label{tab:task}
\end{table}

\begin{figure*}[!t]
    \centering
    \includegraphics[width=0.8\textwidth]{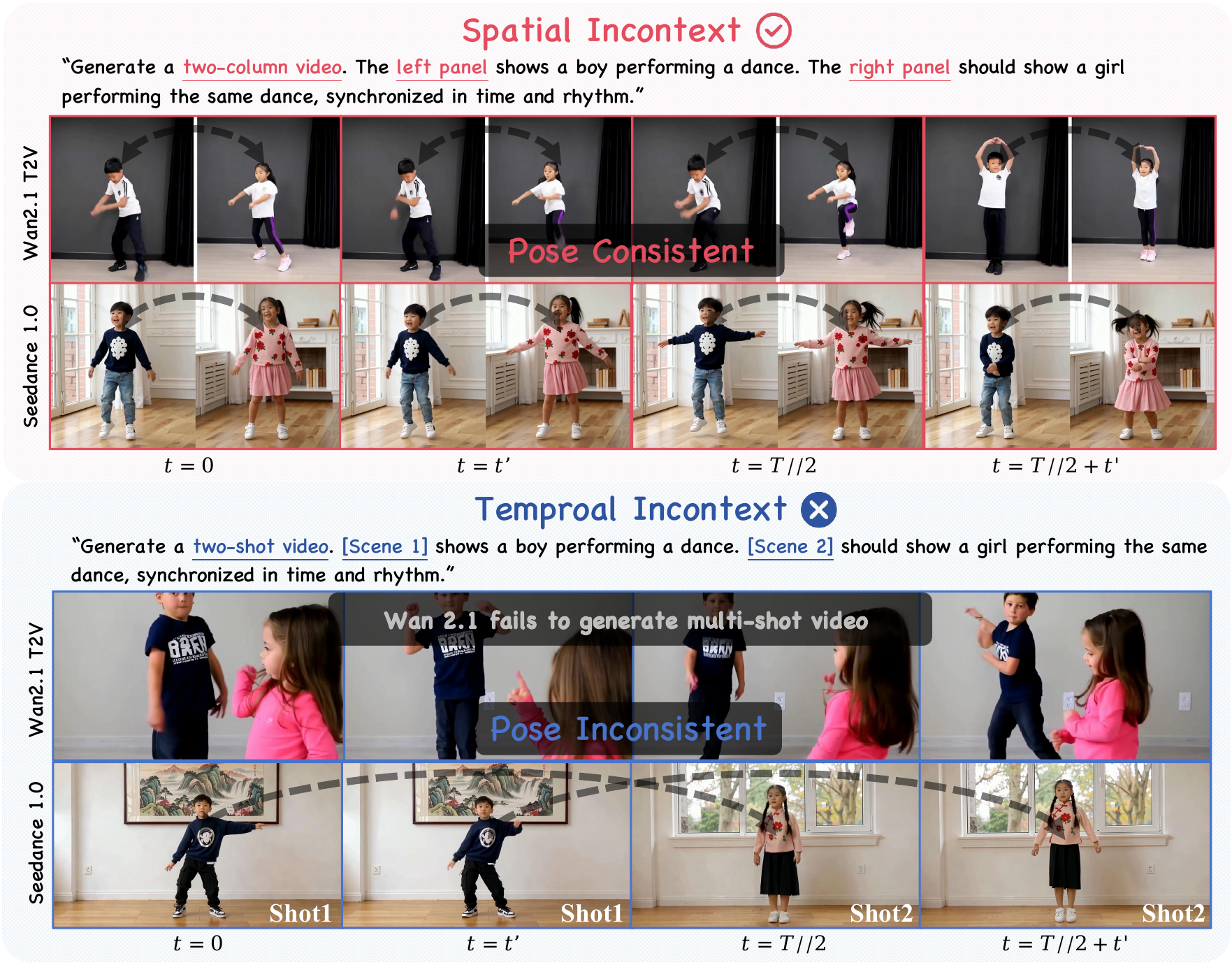}
    \caption{Video diffusion models are inherently capable of handling temporal consistency through spatial context.}
    \label{fig:incontext}
\end{figure*}

In this work, we study both appearance and temporal video reference, across several representative tasks summarized in Table~\ref{tab:task}. To better unify these tasks, we propose OmniTransfer, as illustrated in Fig.~\ref{fig:omnitransfer}. It consists of four components- Reference Latent Construction, Task-aware Positional Bias, Reference-decoupled Causal Learning and Task-adaptive Multimodal Alignment—described in detail below.

\subsection{Reference Latent Construction}
To handle practical scenarios where reference and target videos may have different resolutions, we construct separate latents for the two inputs (Fig. \ref{fig:omnitransfer}). For the target video, the latent is constructed in the same manner as in Section~\ref{sec:preliminary}, i.e., $l_{\text{tgt}} \in \mathbb{R}^{f\times h_{\text{tgt}} \times w_{\text{tgt}} \times (2n+4)} = [c_{\text{tgt}}, m_{\text{tgt}}, z^t_{\text{tgt}}]$. For the reference video, the latent representation is defined as: $l_{\text{ref}}\in \mathbb{R}^{f\times h_{\text{ref}} \times w_{\text{ref}} \times (2n+4)} = [c_{\text{ref}}, m_{\text{ref}}, z^0_{\text{ref}}]$. The conditional latent $c_{\text{ref}}$ corresponds to the feature encoded by the VAE. The mask latent $m_{\text{ref}}$ is assigned task-specific flags: $-1$ for temporal tasks, $-2$ for ID transfer, and $-3$ for style transfer. Notably, the reference latent $z^0_{\text{ref}}$ is kept noise-free to maximally preserve its information.

\subsection{Task-aware Positional Bias}
IC-Lora~\cite{lhhuang2024iclora} proposes the assumption that \textit{``text-to-image models inherently possess in-context generation capabilities''}.
This hypothesis has been well validated in image customization methods~\cite{dreamo,xiao2025omnigen,chen2025xverse}, and is also supported in appearance-consistent video customization~\cite{consistid,standin,liu2025phantom}, where reference appearances are leveraged through temporal in-context learning.
However, it remains unclear whether current video diffusion models exhibit comparable in-context capabilities for temporal consistency tasks.
To investigate this, we directly evaluate the video consistency of two representative models, Wan 2.1~\cite{wan2025} and Seedance~\cite{gao2025seedance}, under the T2V setting, as shown in Fig.~\ref{fig:incontext}. 
We observe that when generating side-by-side videos, both models easily maintain motion consistency across columns. 
In contrast, when generating two temporally consecutive shots, the model fails to keep actions consistent across two shots.
This observation motivates us to propose a new assumption for video models: \textbf{video diffusion models are inherently capable of handling temporal consistency through spatial context.}


Based on the proposed assumption, we introduce Task-aware Positional Bias. Specifically, for temporal reference tasks, we add an offset to the RoPE of the reference video along the spatial (width) dimension, aiming to leverage spatial in-context cues to enhance temporal consistency. The offset is set to the width of the target video, $w_{\text{tgt}}$.
In contrast, for appearance reference tasks, to exploit the temporal propagation of appearance information in the video model, we add an offset along the temporal dimension, equal to the number of frames in the target video, $f$. In summary, we define the RoPE rotation of the reference latents as $R_\theta^{*}$:
\begin{equation}
R_\theta^{*}(\cdot) \!=  \!
\begin{cases} 
R_\theta(\cdot, \Delta\!=\!(0, w_{\text{tgt}}, 0)), \!\!\!\!& \!\text{for temporal ref.}\\
R_\theta(\cdot, \Delta\!=\!(f, 0, 0)), \!\!\!\!& \!\text{for appearance. ref.},
\end{cases} 
\label{eq:TPB}
\end{equation}
where $\Delta = (\Delta_T, \Delta_W, \Delta_H)$ represents the offsets applied along the temporal, width, and height dimensions.

\subsection{Reference-decoupled Causal Learning}
A straightforward way to enable interaction between reference and target videos is through joint self-attention. However, our experiments reveal that this bidirectional attention mechanism may lead to two main issues in video transfer tasks:
1) The generated videos often exhibit a simple “copy-paste” effect from the reference video. We attribute this to the reference branch’s full access to the target video context, which encourages it to adopt a target-like representation, resulting in direct copying of simple patterns.
2) Concatenating the reference and target videos for joint self-attention doubles the number of tokens, leading to a fourfold increase in computational complexity, which is often impractical.

To address the aforementioned issues, we propose fully decoupling the reference and target branches, as illustrated in Fig.~\ref{fig:omnitransfer}.
Formally, both reference and target tokens are first projected into the queries, keys, and values, yielding $Q_{\text{ref}}$, $K_{\text{ref}}$, $V_{\text{ref}}$, $Q_{\text{tgt}}$, $K_{\text{tgt}}$ and $V_{\text{tgt}}$. 
Next, task-aware positional bias is applied to $Q_{\text{ref}}$ and $K_{\text{ref}}$, while $K_{\text{tgt}}$ and $V_{\text{tgt}}$ use the standard RoPE positional encoding.
Subsequently, attention interacts between the two branches in a causal manner.
The reference branch performs intra-branch self-attention to capture internal contextual dependencies:
\begin{equation}
\text{Attn}_{\text{ref}} = \text{Attn}(R^{*}_{\theta}(Q_{\text{ref}}),R^{*}_{\theta}(K_{\text{ref}}),V_{\text{ref}}),
\end{equation}
while the target branch integrates information from both its own features and the reference features by concatenating the keys and values:
\begin{equation}
\text{Attn}_{\text{tgt}} \!\!= \!\!\text{Attn}(\!R_{\theta}(\!Q_{\text{tgt}}\!),\![R_{\theta}(\!K_{\text{tgt}}\!);\!R^{*}_{\theta}(K_{\text{ref}})],\![V_\text{tgt}; \!V_\text{ref}]),
\end{equation}
where $[\cdot;\cdot]$ denotes token-wise concatenation.

We further decouple the time embeddings of the two branches.
Specifically, the reference branch adopts a fixed $t=0$, making it independent of the target video’s noise level.
Thanks to this design, the reference branch becomes time-invariant during inference, effectively reducing computational overhead and shortening the overall generation time by 20\% compared to the standard architecture.

\subsection{Task-adaptive Multimodal Alignment }

In multi-task video transfer, different tasks demand reference information with distinct semantic focuses. 
However, conventional in-context learning in diffusion models primarily captures shallow visual correspondences rather than semantic intent, limiting their adaptability across tasks.
To overcome this, we replace the original T5~\cite{raffel2020exploring} features in Wan with representations from MLLM, i.e., Qwen-2.5-VL~\cite{Qwen2.5-VL}, introducing richer visual-semantic cues that enhance reference understanding and task-level alignment.

The MLLM takes as input the first-frame tokens of the target video, the reference video tokens, template tokens, and prompt tokens. 
To extract task-specific representations, we draw inspiration from MetaQuery~\cite{pan2025transfer} and introduce a set of learnable tokens dedicated to each task. 
For temporal tasks, the MetaQuery aggregates temporal cues from the reference video together with the target’s first-frame content, effectively capturing cross-frame dynamics in the generated sequence. 
For appearance tasks, it instead fuses identity or style information from the reference with the semantic context derived from the prompt tokens.

To preserve the multimodal reasoning capability while enabling parameter-efficient adaptation, the MLLM is fine-tuned using LoRA~\cite{hu2022lora}.
For seamless integration with the diffusion model, its outputs are passed through a three-layer MultiLayer Perceptron (MLP) and injected solely into the target branch, thereby preventing interference with the reference branch.
Consequently, the cross-attention in the target branch is defined as ${\text{Attn}(Q_{\text{tgt}}, K_\text{MLLM}, V_\text{MLLM})}$, where $K_\text{MLLM}$ and $V_\text{MLLM}$ denote the keys and values derived from the aligned MLLM features.

\section{Experiment}

\begin{figure*}[!t]
    \centering
    \includegraphics[width=\textwidth]{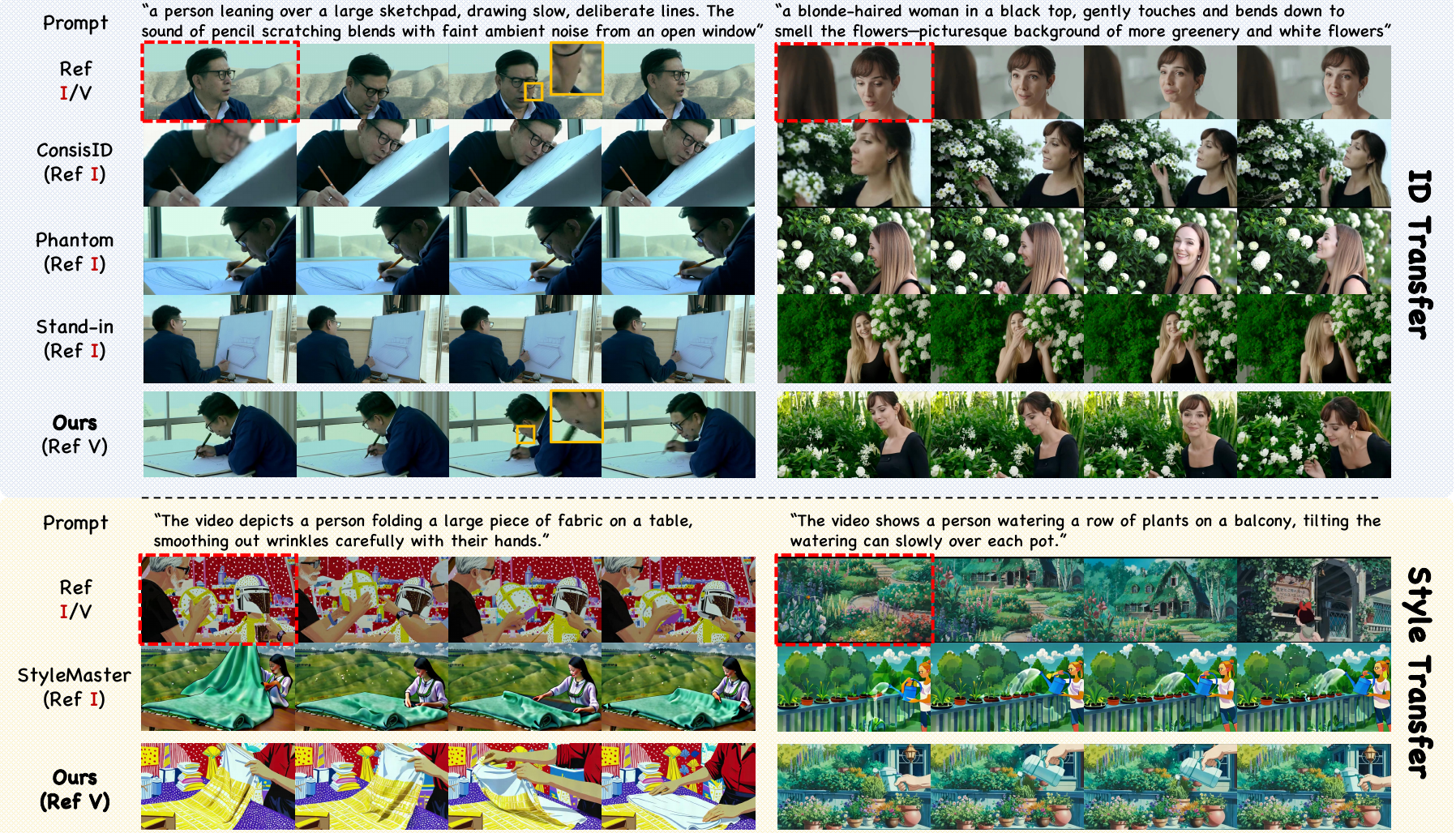}
    \caption{Qualitative comparison for appearance video transfer. The \textcolor{yellow}{yellow box} highlights how OmniTransfer captures richer appearance details from multiple cross-view video frames. The \textcolor{red}{red box} denotes the input image for image-reference methods.}
    \label{fig:spatial_compare}
\end{figure*} 

\begin{figure*}[!t]
    \centering
    \includegraphics[width=\textwidth]{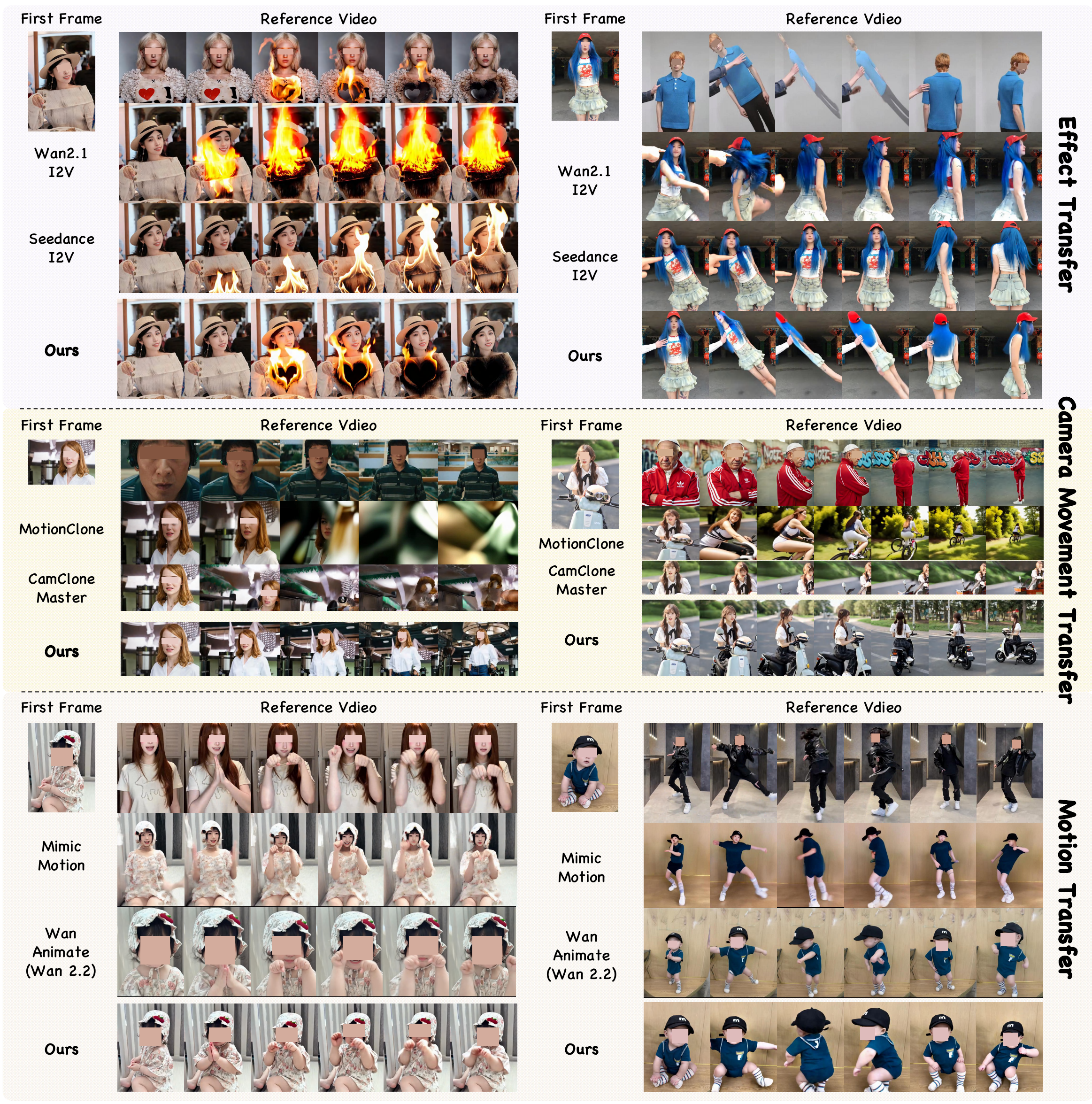}
    \caption{Qualitative comparison for appearance video transfer.}
    \label{fig:temporal_compare}
\end{figure*}

\subsection{Implementation Details}
Our training process is divided into three sequential stages with distinct optimization objectives.
In the first stage, we train the DiT blocks via in-context learning. Subsequently, we freeze the DiT blocks and focus on training the connector to align the MLLM with diffusion models. Finally, we unfreeze all components to conduct joint fine-tuning.
In terms of training hyperparameters, we set the learning rate to $1e-5$, with a batch size of 16. The three stages are trained for 10,000, 2,000, and 5,000 training steps, respectively.

Due to the lack of public data sets containing reference video pairs at present, we collected our own data sets from the Internet to support spatio-temporal video transfer. 

\subsection{Evaluation Details}
To evaluate spatio-temporal video transfer, we curated dedicated test sets for each subtask. For ID transfer, 50 videos of diverse individuals are paired with two prompts each to test identity consistency. Style transfer includes 20 unseen visual styles, each with two prompts to assess stylistic variation. Effect transfer contains 50 unseen effects from visual-effects websites. Camera movement transfer uses 50 professionally shot videos with complex trajectories. Motion transfer comprises 50 popular dance videos covering diverse dynamic and fine-grained body motions.

\subsection{Comparison}
\subsubsection{ID Transfer}
We compared our method with State-Of-The-Art (SOTA) video ID preservation approaches, including ConsisID~\cite{consistid}, Phantom~\cite{liu2025phantom} and Stand-in~\cite{standin}. Following Phantom, we measure text–video alignment using the CLIP-T~\cite{wang2022internvideo} score, and assess identity consistency with several face recognition models, including Sim-Arc~\cite{Deng_2019_CVPR}, Sim-Cur~\cite{Huang_2020_CVPR}, and Sim-Glint~\cite{an_2021_pfc_iccvw}. We further extend these metrics to video-level similarity by matching four evenly sampled reference frames with all generated frames, denoted as VSim-Arc, VSim-Cur, and VSim-Glint.

\setlength{\tabcolsep}{7mm}  
\begin{table}[t]

\centering
\renewcommand{\arraystretch}{1}
\begin{tabular}{c|ccc|c}  
\Xhline{1pt}
Method & VSim-Arc $\uparrow$ & VSim-Cur $\uparrow$ & VSim-Glint $\uparrow$ & CLIP-T $\uparrow$ \\  
\midrule  
ConsisID~\cite{consistid} & 0.34 & 0.32 & 0.36 & \textbf{21.54} \\
Phantom~\cite{liu2025phantom} & 0.45 & 0.41 & 0.47 & 20.34 \\
Stand-in~\cite{standin} & 0.30 & 0.21 & 0.26 & 20.38 \\
\textbf{Ours} & \textbf{0.48} & \textbf{0.43} & \textbf{0.51} & 20.35 \\ 
\Xhline{1pt}
\end{tabular}  
\caption{ Quantitative comparison for ID Transfer.}
\label{tab: ID_trans}
\end{table}

Leveraging multi-view facial information, our method achieves high facial similarity with natural, fluid motions. As shown in the yellow box of Fig.~\ref{fig:spatial_compare}, it preserves fine details like acne across frames, which is difficult for image reference methods. In Fig.~\ref{fig:spatial_compare} (right), our method generates diverse poses—including frontal, profile, and tilted views—while maintaining high facial similarity, highlighting the advantage of using video as the reference.

\subsubsection{Style Transfer}
We compared our method with SOTA text-to-video stylization approaches, StyleCrafter~\cite{stylecrafter} and StyleMaster~\cite{stylemaster}. Following StyleMaster, we evaluated text–video alignment with CLIP-T, while aesthetics with the Aesthetics Score~\cite{huang2023vbench}, and style consistency with the video CSD Score~\cite{somepalli2024measuring} (VCSD) using four sampled reference frames.

\setlength{\tabcolsep}{11.5mm}  
\begin{table}[t]

\centering
\renewcommand{\arraystretch}{1} 
\begin{tabular}{c|ccc}  
\Xhline{1pt}
Method & VCSD $\uparrow$ & CLIP-T $\uparrow$ & Aesthetics $\uparrow$ \\  
\midrule  
StyleCrafter~\cite{stylecrafter} & 0.44 & 24.72  & 0.47 \\
StyleMaster~\cite{stylemaster} & 0.29 & 26.82 & 0.59 \\
\textbf{Ours} & \textbf{0.51} & \textbf{27.16} & \textbf{0.61} \\ 
\Xhline{1pt}
\end{tabular}  
\caption{ Quantitative comparison for Style Transfer.}
\label{tab: StyleTransfer}
\end{table}

Table~\ref{tab: StyleTransfer} shows that our method outperforms others on all three metrics. For qualitative comparison in Fig.~\ref{fig:spatial_compare}, we only include StyleMaster, as StyleCrafter yields lower visual quality due to its earlier UNet-based design. The results demonstrate that our method effectively captures style from the video, surpassing previous image-based methods.

\subsubsection{Effect Transfer}
\label{effect_comparison}
Since our test set comes from effect websites with effects that existing commercial LoRA models cannot reproduce, we instead compare our method with SOTA image-to-video models, Wan 2.1 I2V~\cite{wan2025} and Seedance~\cite{gao2025seedance}, using prompts generated by Qwen-2.5-VL~\cite{Qwen2.5-VL} from the reference effect videos. 
With no standard metrics for effect consistency, we conduct a user study with 20 volunteers, who rate effect fidelity, first-frame consistency, and overall visual quality on a five-point scale.

As shown in Table~\ref{tab: Effect Transfer}, our method achieves the highest scores on all three metrics. 
Qualitative results in Fig.~\ref{tab: Effect Transfer} further show that only our method accurately reproduces the effects of the reference videos, outperforming Seedance and Wan I2V.
This demonstrates that text alone is insufficient, emphasizing the value of temporal video references.

\setlength{\tabcolsep}{8mm}  
\begin{table}[t]

\centering
\renewcommand{\arraystretch}{1}
\begin{tabular}{c|ccc}  
\Xhline{1pt}
Method & Effect Fidelity $\uparrow$ & Image Consistency $\uparrow$ & Quality $\uparrow$ \\  
\midrule  
Wan2.1 I2V~\cite{wan2025} & 1.81 & 2.89 & 2.03 \\
Seedance I2V~\cite{gao2025seedance} & 1.95 & 3.20 & 2.42 \\
\textbf{Ours} & \textbf{3.45} & \textbf{3.49} & \textbf{3.27} \\ 
\Xhline{1pt}
\end{tabular}  
\caption{ User study on Effect Transfer.}
\label{tab: Effect Transfer}
\end{table}

\setlength{\tabcolsep}{7mm}  
\begin{table}[t]

\centering
\renewcommand{\arraystretch}{}
 
\begin{tabular}{c|ccc}  
\Xhline{1pt}
Method & Camera Fidelity $\uparrow$ & Image Consistency $\uparrow$ & Quality $\uparrow$ \\  
\midrule  
MotionClone~\cite{motionclone} & 1.75 & 1.23 & 1.29 \\
CamCloneMaster~\cite{camclonemaster} & 1.79 & 1.45 &  1.29 \\
\textbf{Ours} & \textbf{4.19} & \textbf{3.89} & \textbf{3.85} \\ 
\Xhline{1pt}
\end{tabular}  

\caption{ User study on Camera Movement Transfer.}
\label{tab: Camera Transfer}
\end{table}

\subsubsection{Camera Movement Transfer}

We compare our method with SOTA camera movement models, MotionClone~\cite{motionclone} and CamCloneMaster~\cite{camclonemaster}. 
Since the estimation of complex camera trajectory remains a challenging problem, we conducted a user study using the same setup as in Section~\ref{effect_comparison}, evaluating camera fidelity, image consistency, and overall quality.

\begin{figure*}[!t]
    \centering
    \includegraphics[width=\textwidth]{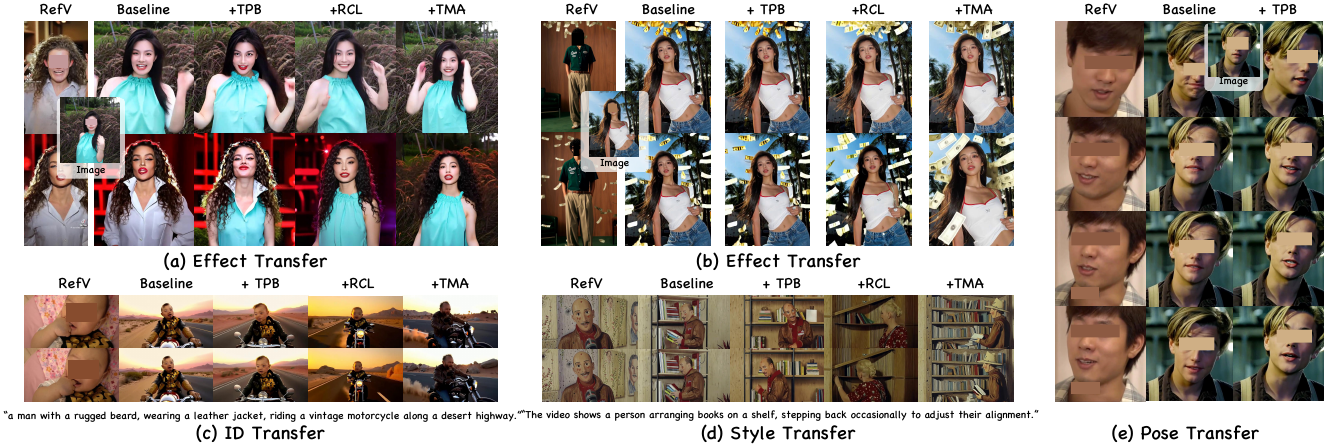}
    \caption{Qualitative ablation study comparison. Best viewed in zoom.}
    \label{fig:ablation}
\end{figure*}

As shown in Table~\ref{tab: Camera Transfer}, our method outperforms all others across the three metrics. The qualitative results in Fig.~\ref{fig:temporal_compare} show that competing methods produce only fixed-resolution outputs, resulting in unavoidable resizing or cropping. Moreover, our approach is the only one that successfully replicates camera movements from cinematic scenes (left) and complex professional tracking shots (right), demonstrating strong generalization capability.

\subsubsection{Motion Transfer}
We compare our method with state-of-the-art motion transfer approaches, MimicMotion~\cite{mimicmotion} and WanAnimate~\cite{wananimate}. As the test set contains in-the-wild videos without paired ground truth, we conducted a user study following  Section~\ref{effect_comparison}, evaluating motion fidelity, image consistency, and overall quality. As shown in Table~\ref{tab: Camera Transfer}, despite using the smaller 14B Wan 2.1 model without additional pose input, our method achieves the highest image consistency, with motion consistency and quality comparable to WanAnimate, which relies on the larger 28B Wan 2.2 model.

Qualitative comparisons in Fig.~\ref{tab: MotionTransfer} show that baseline methods need to align the first frame to a target pose, often causing appearance loss (bottom-right of Fig.~\ref{tab: MotionTransfer}) and unnatural motion. In contrast, our method requires no pose guidance, preserves appearance, produces natural motions, and easily extends to multi-person scenarios (Fig.~\ref{fig:abstract}).

\setlength{\tabcolsep}{8 mm}  
\begin{table}[t]
\centering
\renewcommand{\arraystretch}{1}
\begin{tabular}{c|ccc}  
\Xhline{1pt}
Method & Motion Fidelity $\uparrow$ & Image Consistency $\uparrow$ & Quality $\uparrow$ \\  
\midrule  
MimicMotion~\cite{mimicmotion} & 2.67 & 1.84 & 2.02 \\
WanAnimate~\cite{wananimate} & \textbf{3.71} & 3.53 & \textbf{3.48} \\
\textbf{Ours} & 3.62 & \textbf{3.88} & 3.45  \\ 
\Xhline{1pt}
\end{tabular}  
\caption{ User study on Motion Transfer.}
\label{tab: MotionTransfer}
\end{table}

\subsection{Ablation Study}
Ablation experiments are conducted incrementally from a Baseline by successively adding TPB, RCL and TMA:

\noindent\textbf{Baseline. } We use a vanilla in-context learning setup, where reference features are concatenated along the temporal dimension~\cite{liu2025phantom, camclonemaster}, and full attention is applied.

\noindent\textbf{+TPB. } This model incorporates the  Task-aware Positional Bias (TPB) defined in Eq.~(\ref{eq:TPB}).

\noindent\textbf{+RCL. } Full attention is replaced by Reference-decoupled Causal Learning (RCL).

\noindent\textbf{+TMA (Full Model). } The Task-adaptive Multimodal Alignment (TMA) is added to form the complete model.

The ablation study is conducted on 20 cases for each of the appearance and temporal transfer tasks, with a user study rating overall quality and reference consistency on a five-point scale. The results are presented in Table~\ref{tab:user_study}, with a qualitative comparison in Fig.~\ref{fig:ablation} for a more comprehensive analysis of each module's contributions.
(1) Baseline: Without TPB, subtle motion cannot be effectively transferred (Fig.~\ref{fig:ablation}-e), and task confusion occurs where appearance cues leak into temporal transfer (Fig.~\ref{fig:ablation}-a). 
(2) +TPB: Fine-grained motion transfer is achieved by leveraging spatial context (Fig.~\ref{fig:ablation}-e), while task confusion is alleviated by different RoPE biases (Fig.~\ref{fig:ablation}-a).
(3) +RCL: The copy–paste issue is alleviated. For example, in Fig.~\ref{fig:ablation}-a and d, the identity is not fully copied; in Fig.~\ref{fig:ablation}-c the face appears more natural, highlighting the effectiveness of causal attention. Additionally, RCL also improves inference speed by 20\%.
(4) +TMA (Full Model): The TMA module substantially enhances semantic understanding. For instance, in Fig.~\ref{fig:ablation}-a, the model understands that the scene remains unchanged; in Fig.~\ref{fig:ablation}-b, it correctly understands and generates the money rather than simply copying reference patterns; and in Fig.~\ref{fig:ablation}-c, richer semantic control enables realistic details such as a leather jacket, beard, and side-face angle. These results demonstrate the effectiveness of semantic guidance in improving scene controllability.

\setlength{\tabcolsep}{3.5mm}{
\begin{table}[t]

\centering
\renewcommand{\arraystretch}{1}
\begin{tabular}{c|ccc|cc|c}
\Xhline{1pt}
\multirow{2}{*}{ Method} &
\multicolumn{3}{c|}{ Components} &
\multicolumn{1}{c}{ Appearance} &
\multicolumn{1}{c|}{ Temporal} &
\multirow{2}{*}{ Time} \\
\cline{2-6}
 & TPB & RCL & TMA &
 Consistency/Quality &  Consistency/Quality & \\
\midrule
Baseline &  &  &  & 2.36 / 2.53 & 2.69 / 2.70 & 180s \\
+TPB & \checkmark &  &  & 2.82 / 2.86 & 2.95 / 2.94 & 180s \\
+RCL & \checkmark & \checkmark &  & 3.10 / 3.16 & 3.13 / 3.10 & \textbf{142s} \\
+TMA (Full) & \checkmark & \checkmark & \checkmark & \textbf{3.27} / \textbf{3.56} & \textbf{3.36} / \textbf{3.51} & 145s \\
\Xhline{1pt}
\end{tabular}

\caption{ Ablation study. “Time” indicates inference time per sample (480p, 81 frames, 8×NVIDIA A100).}
\label{tab:user_study}
\end{table}
}

\section{Conclusion}
In this work, we introduced OmniTransfer, a unified framework for spatio-temporal video transfer. By integrating Task-aware Positional Bias, Reference-decoupled Causal Learning, and Task-adaptive Multimodal Alignment, our approach effectively leverages multi-view and temporal information from reference videos, enabling fine-grained and consistent video generation across diverse tasks. Extensive experiments verify that OmniTransfer not only achieves superior performance in both appearance and temporal transfer but also establishes a new paradigm for flexible, high-fidelity video generation.

\section{Acknowledgments}
We would like to express our sincere gratitude to Junjie Luo, Pengqi Tu, Qi Chen, Qichao Sun and Wanquan Feng for their insightful discussions and valuable data contributions.

\bibliographystyle{ieeenat_fullname}
\bibliography{main}

\clearpage
\appendix
\section{Additional Comparison Results} 
\label{sec: additional comparisons}
\begin{figure*}[!t]
    \centering
    \includegraphics[width=\textwidth]{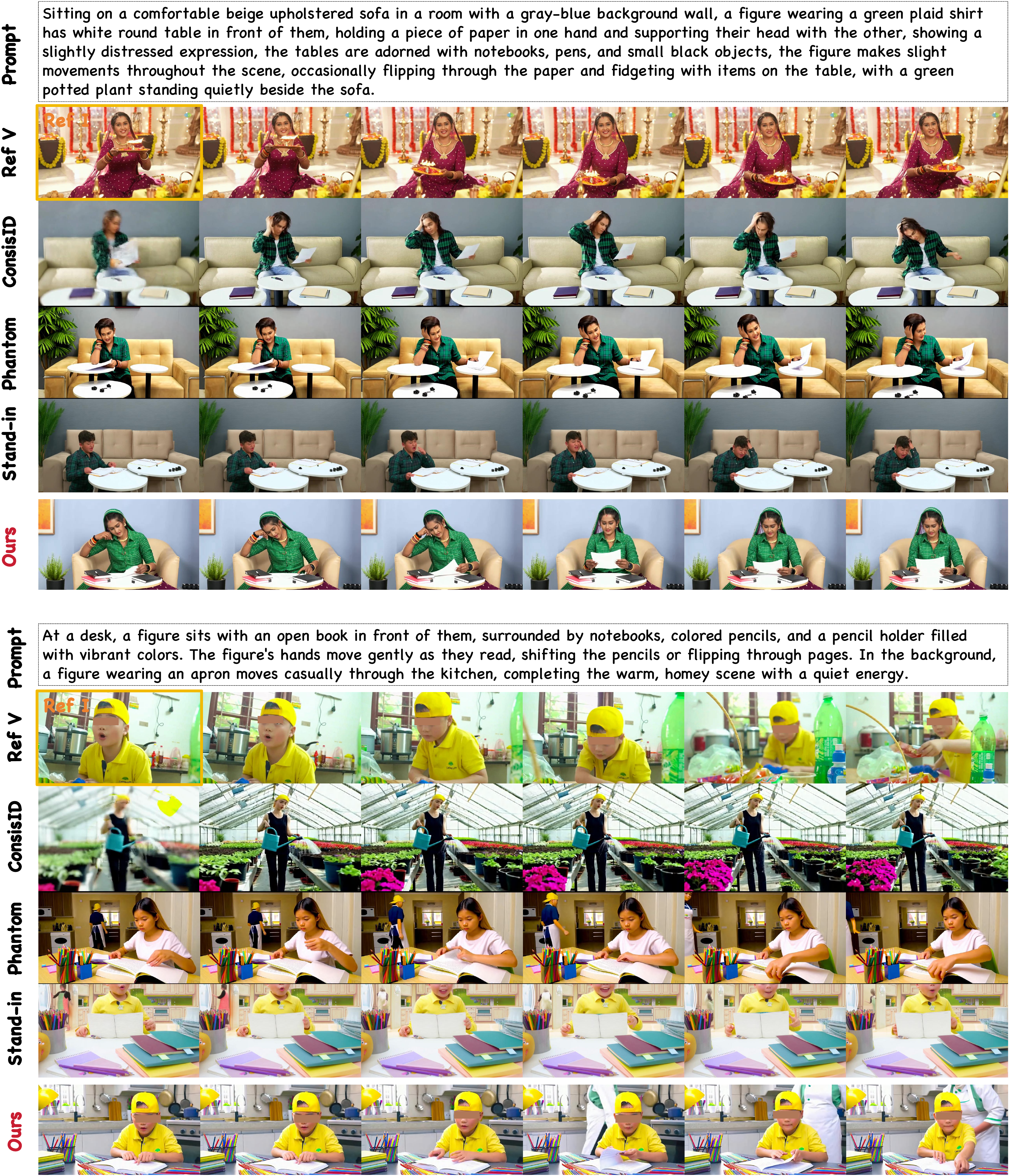}
    \caption{Additional ID transfer comparison with ConsisID~\cite{consistid}, Phantom~\cite{liu2025phantom}, and Stand-in~\cite{standin} (Set 1).}
    \label{fig:id_s1}
\end{figure*}

\begin{figure*}[!t]
    \centering
    \includegraphics[width=\textwidth]{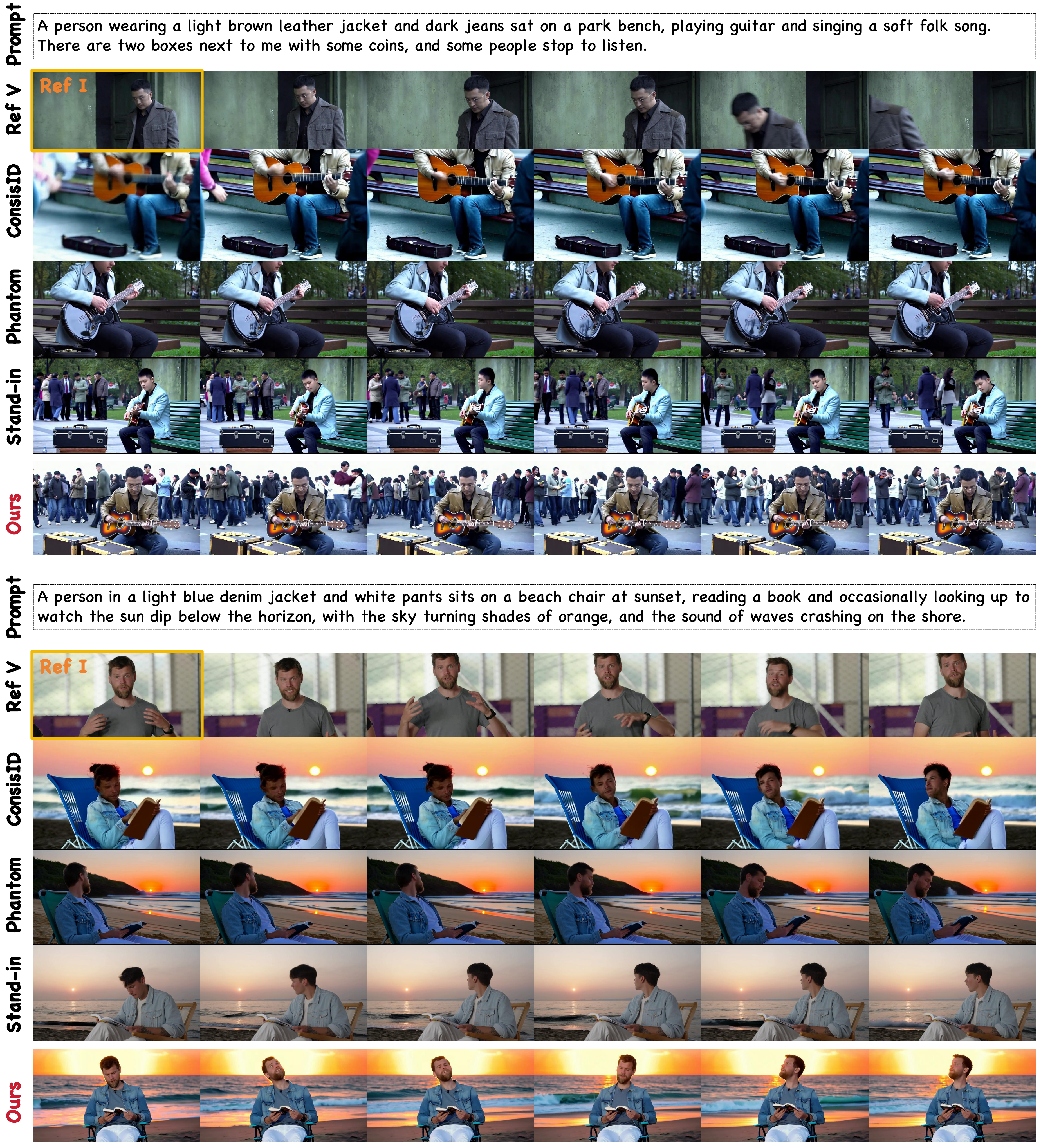}
    \caption{Additional ID transfer comparison with ConsisID~\cite{consistid}, Phantom~\cite{liu2025phantom}, and Stand-in~\cite{standin} (Set 2).}
    \label{fig:id_s2}
\end{figure*}

\begin{figure*}[!t]
    \centering
    \includegraphics[width=\textwidth]{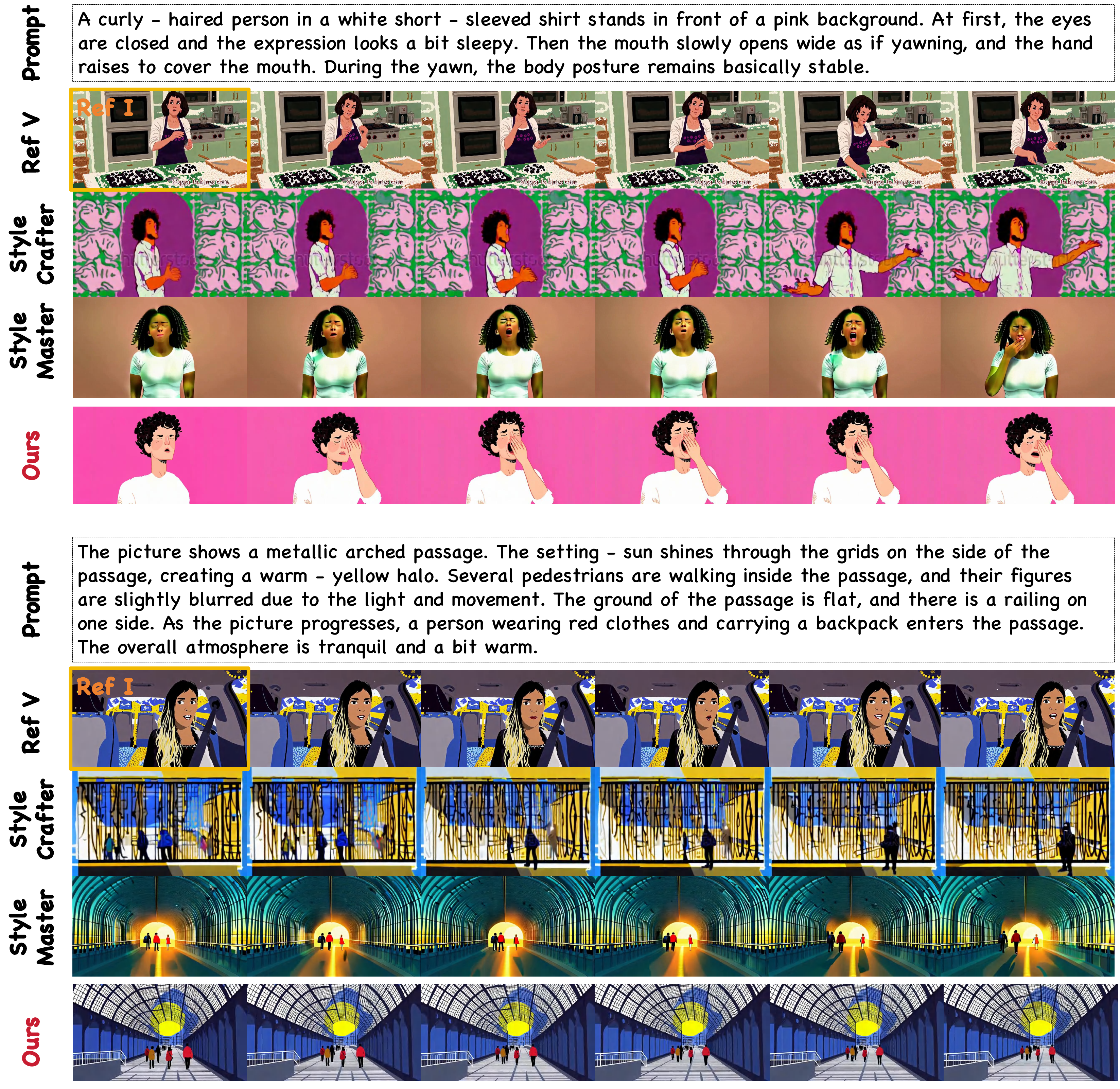}
    \caption{Additional style transfer comparison with StyleCrafter~\cite{stylecrafter} and StyleMaster~\cite{stylemaster} (Set 1).}
    \label{fig:style_s1}
\end{figure*}

\begin{figure*}[!t]
    \centering
    \includegraphics[width=\textwidth]{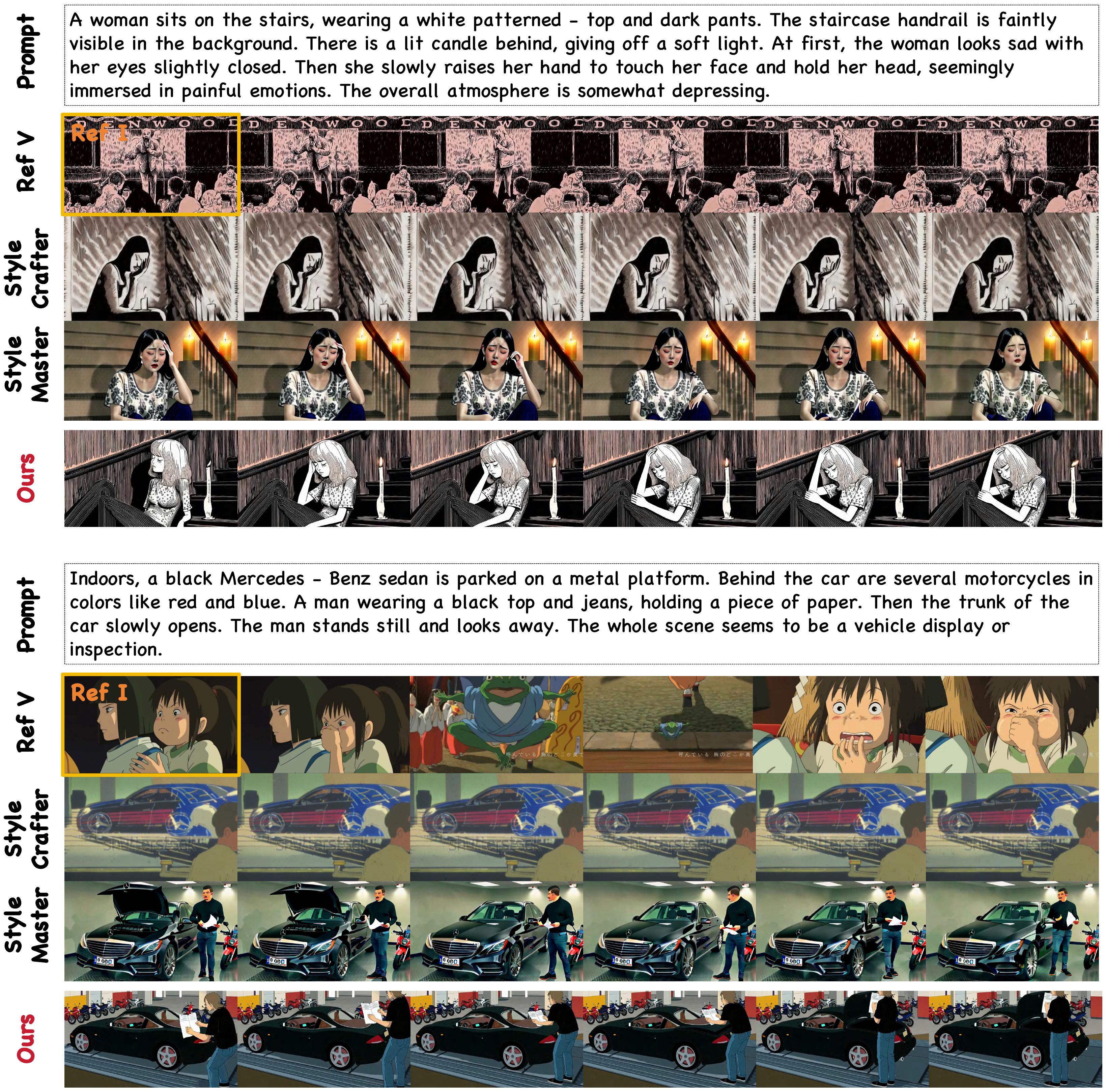}
    \caption{Additional style transfer comparison with StyleCrafter~\cite{stylecrafter} and StyleMaster~\cite{stylemaster} (Set 2).}
    \label{fig:style_s2}
\end{figure*}

\begin{figure*}[!t]
    \centering
    \includegraphics[width=\textwidth]{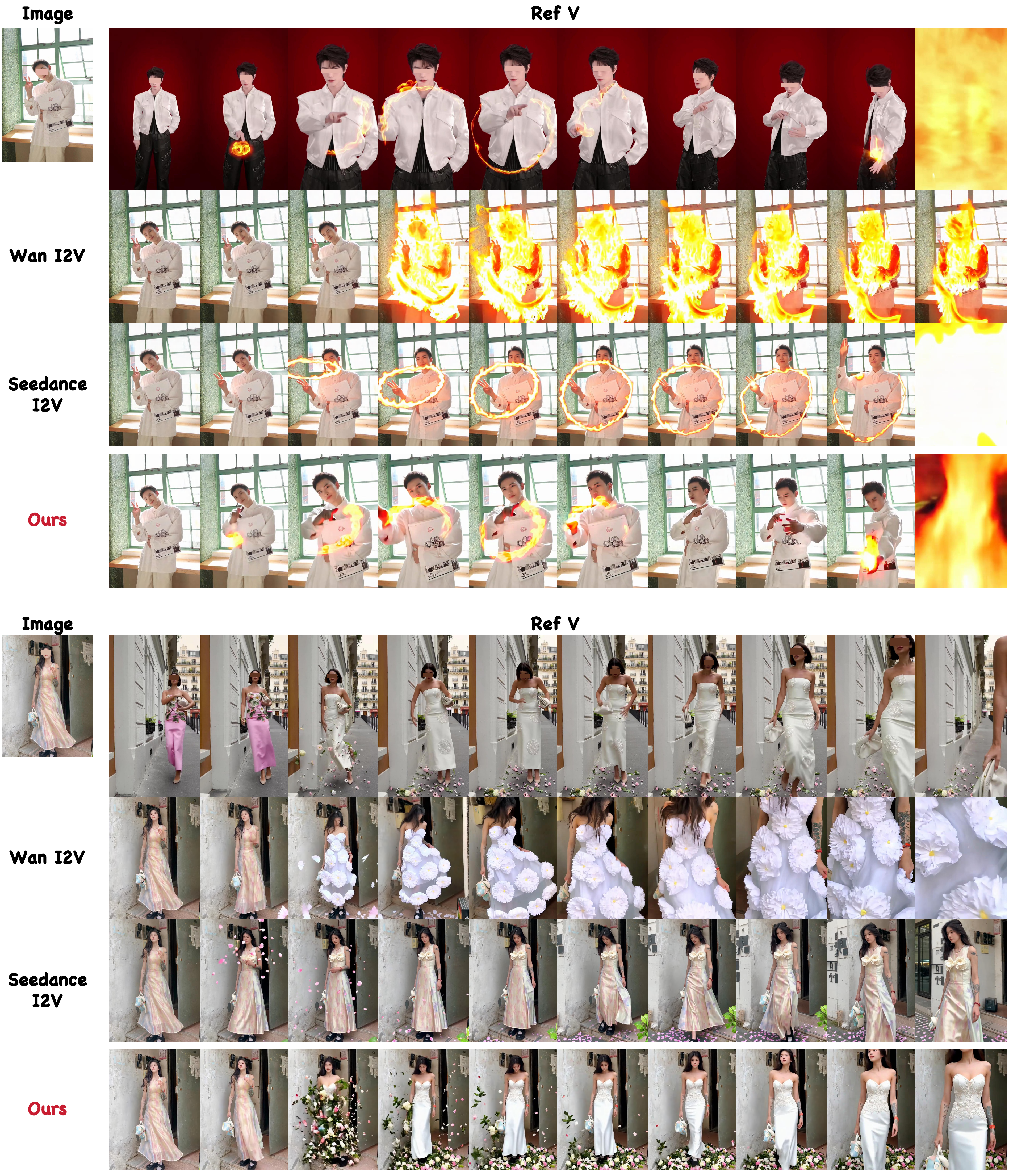}
    \caption{Additional effect transfer comparison with Wan2.1 I2V~\cite{wan2025} and Seedance 1.0 I2V~\cite{gao2025seedance} (Set 1).}
    \label{fig:effect_s1}
\end{figure*}

\begin{figure*}[!t]
    \centering
    \includegraphics[width=\textwidth]{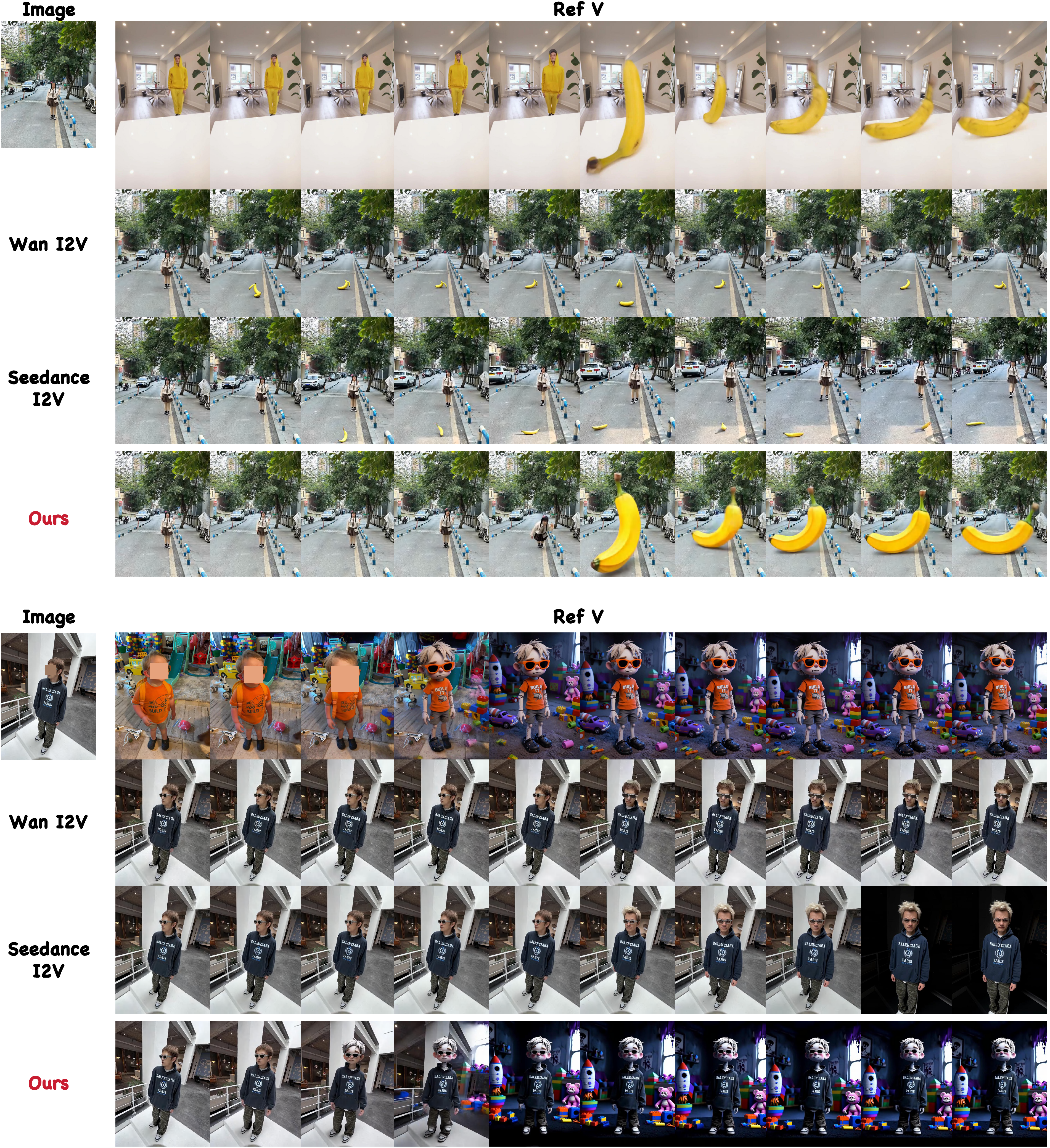}
    \caption{Additional effect transfer comparison with Wan2.1 I2V~\cite{wan2025} and Seedance 1.0 I2V~\cite{gao2025seedance} (Set 2).}
    \label{fig:effect_s2}
\end{figure*}

\begin{figure*}[!t]
    \centering
    \includegraphics[width=\textwidth]{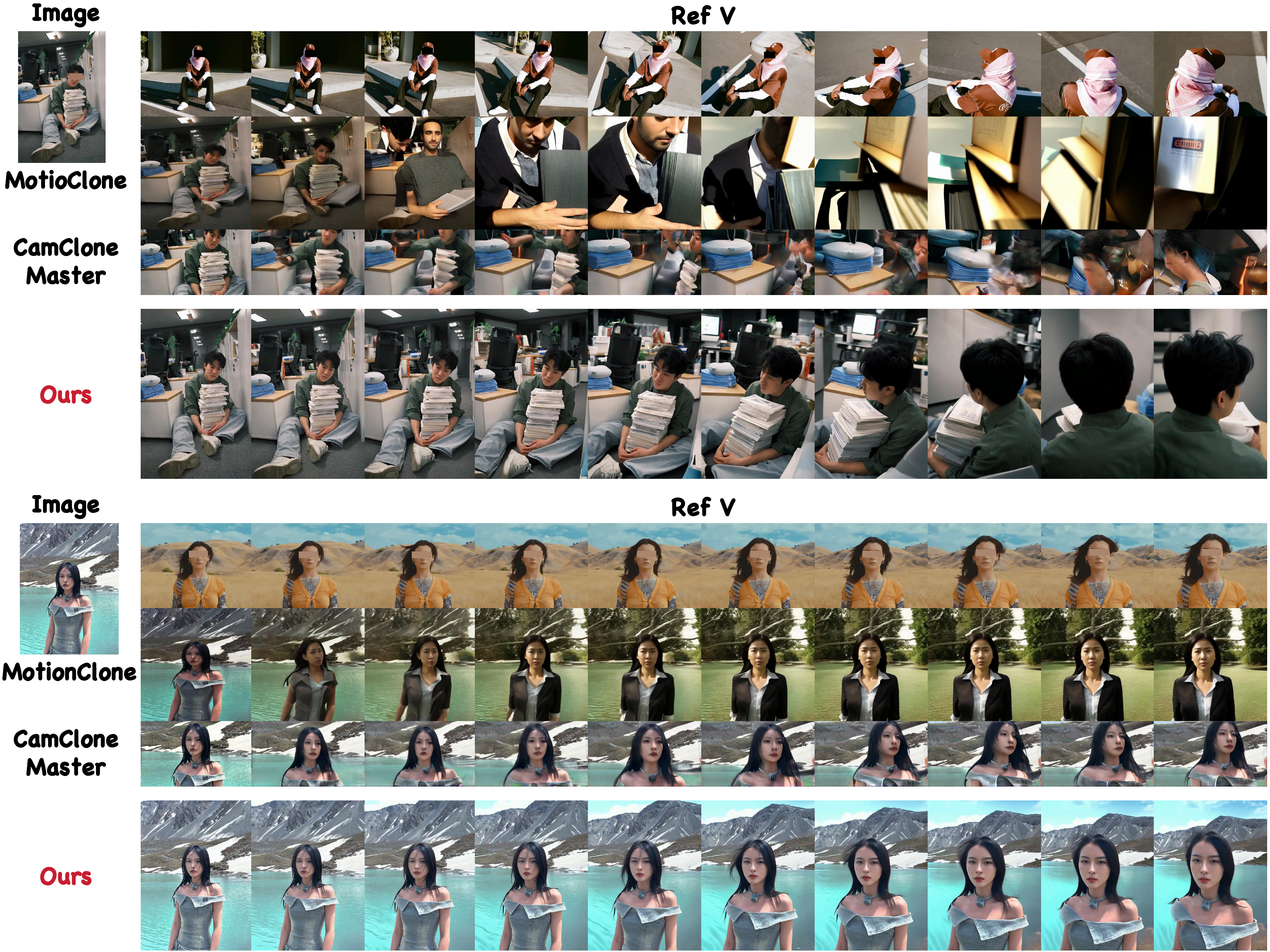}
    \caption{Additional camera movement transfer comparison with MotionClone~\cite{motionclone} and CamCloneMaster~\cite{camclonemaster} (Set 1).}
    \label{fig:camera_s1}
\end{figure*}

\begin{figure*}[!t]
    \centering
    \includegraphics[width=\textwidth]{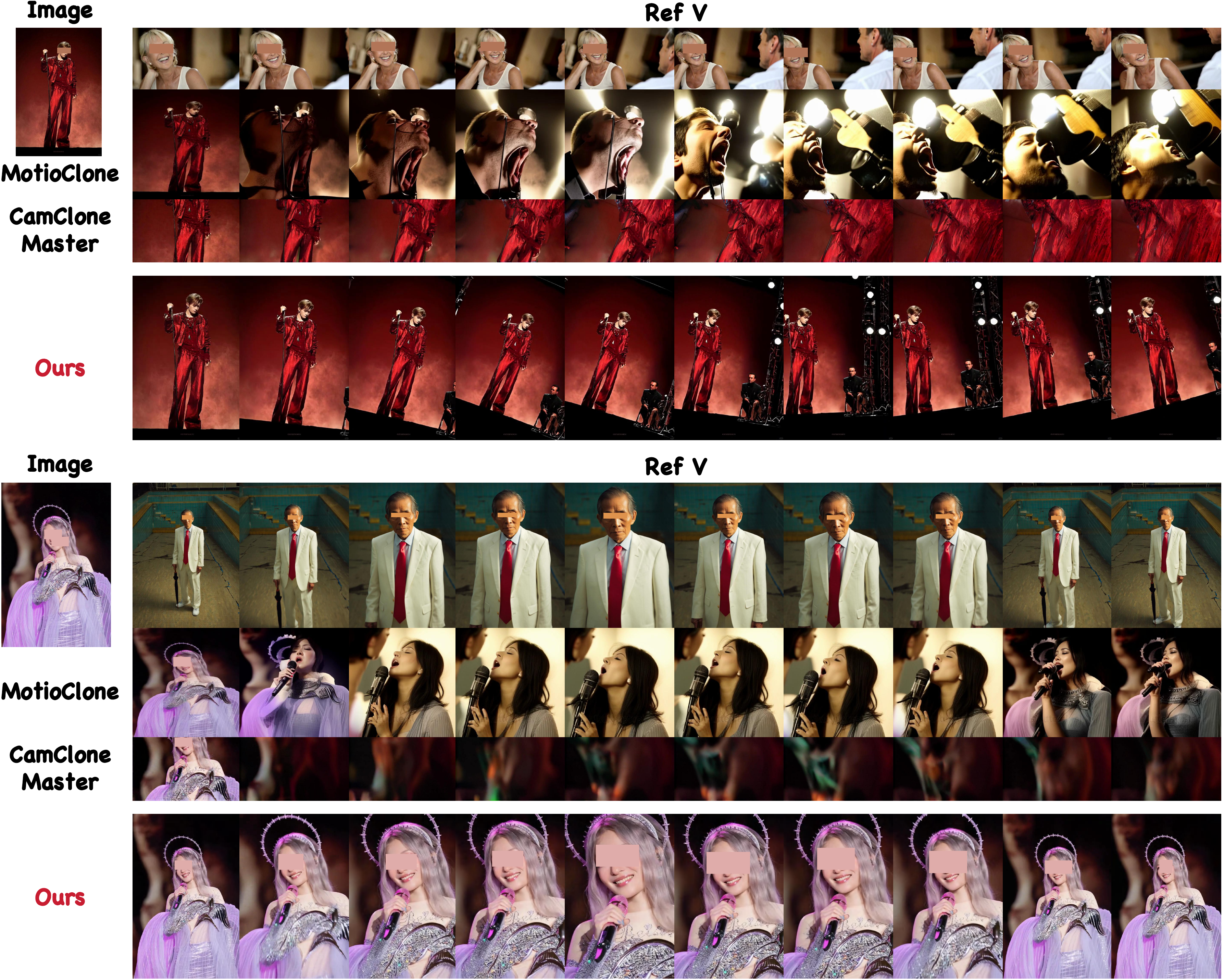}
    \caption{Additional camera movement transfer comparison with MotionClone~\cite{motionclone} and CamCloneMaster~\cite{camclonemaster} (Set 2).}
    \label{fig:camera_s2}
\end{figure*}

\begin{figure*}[!t]
    \centering
    \includegraphics[width=\textwidth]{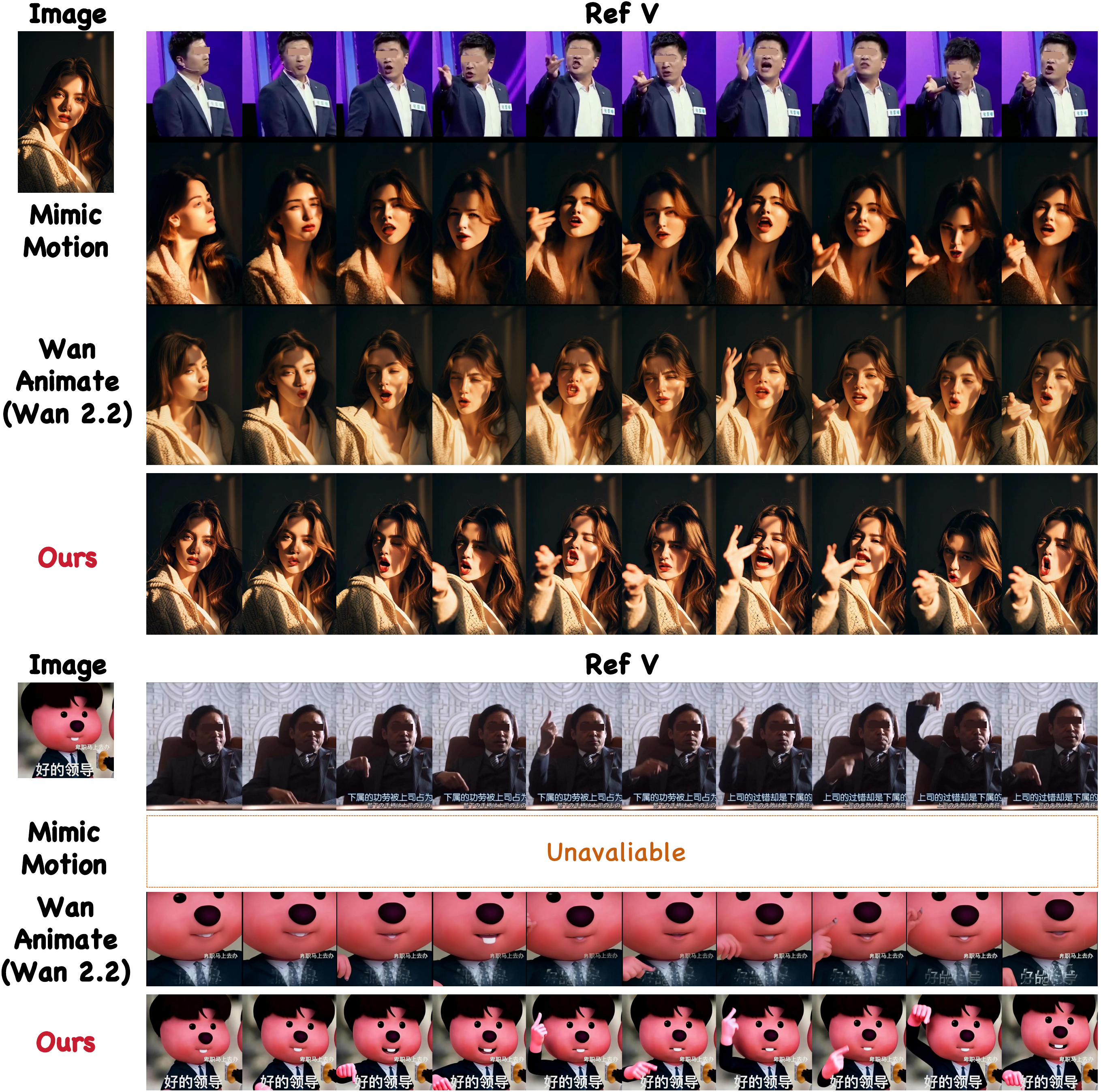}
    \caption{Additional motion transfer comparison with MimicMotion~\cite{mimicmotion} and WanAnimate~\cite{wananimate} (Set 1). Unavailable indicates that pose-based methods fail to generate results due to errors in pose extraction or pose alignment.}
    \label{fig:motion_s1}
\end{figure*}

\begin{figure*}[!t]
    \centering
    \includegraphics[width=\textwidth]{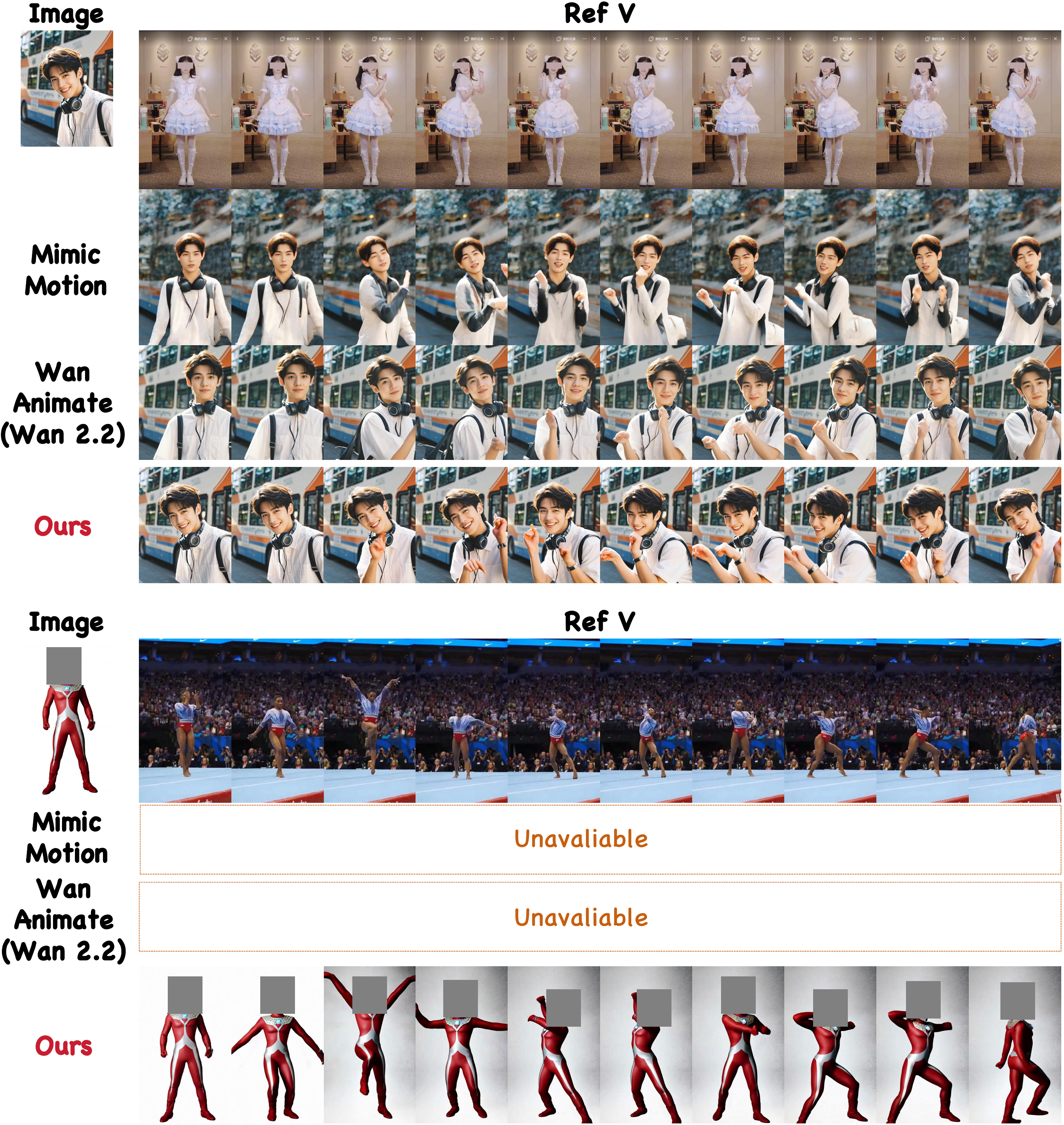}
    \caption{Additional motion transfer comparison with MimicMotion~\cite{mimicmotion} and WanAnimate~\cite{wananimate} (Set 2). Unavailable indicates that pose-based methods fail to generate results due to errors in pose extraction or pose alignment.}
    \label{fig:motion_s2}
\end{figure*}

We present additional comparisons for the ID transfer task in Fig.~\ref{fig:id_s1} and Fig.~\ref{fig:id_s2}. Fig.~\ref{fig:style_s1} and Fig.~\ref{fig:style_s2} show further comparisons for the style transfer task. Fig.~\ref{fig:effect_s1} and Fig.~\ref{fig:effect_s2} provide more comparisons for effect transfer. Fig.~\ref{fig:camera_s1} and Fig.~\ref{fig:camera_s2} present additional results for camera motion transfer, and Fig.~\ref{fig:motion_s1} and Fig.~\ref{fig:motion_s2} show further comparisons for motion transfer.

\section{Video Transfer Combination}
\label{sec: combination}
\begin{figure*}[!t]
    \centering
    \includegraphics[width=\textwidth]{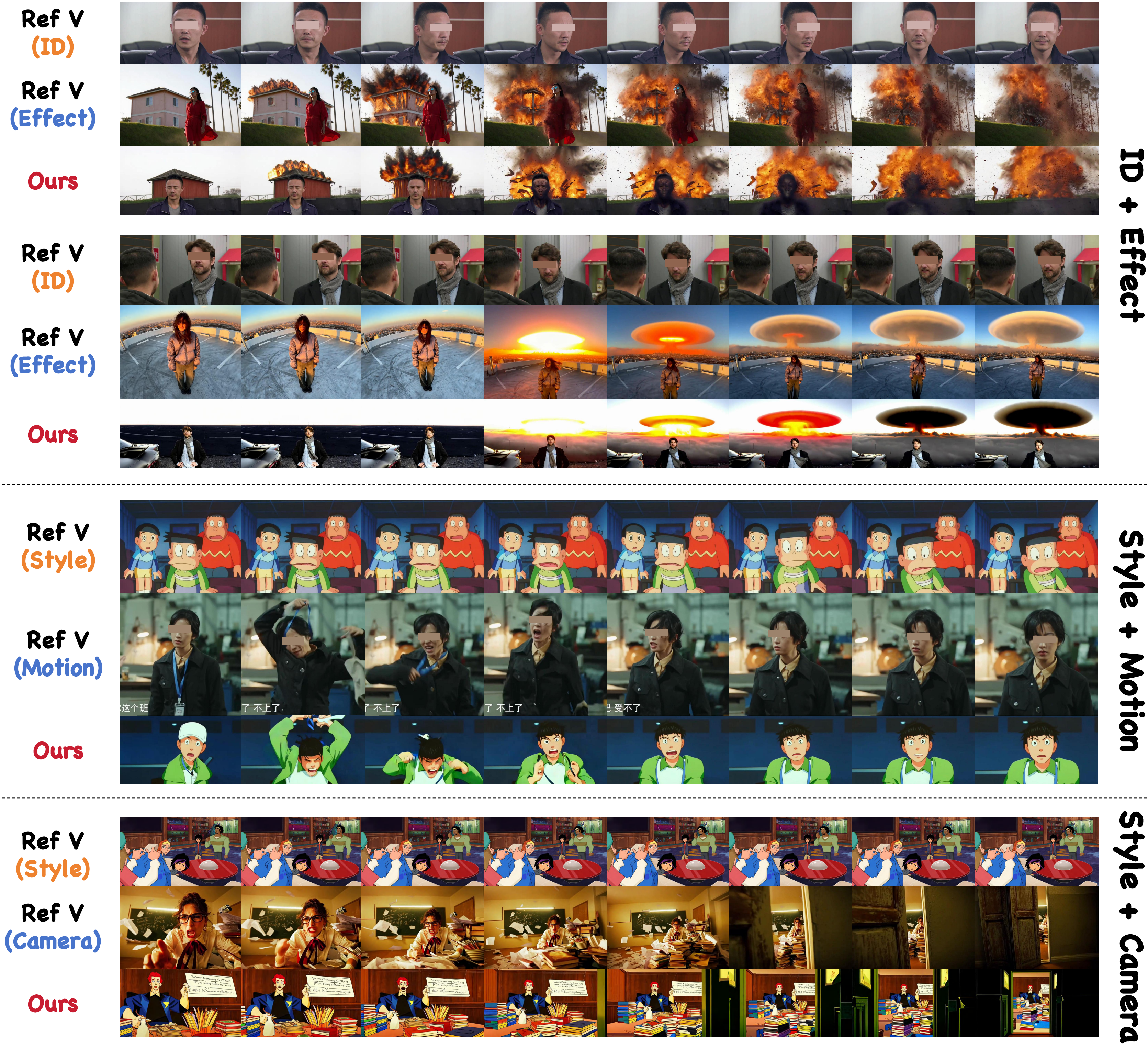}
    \caption{Video transfer combination results(Set 1).}
    \label{fig:comb-s1}
\end{figure*}

\begin{figure*}[!t]
    \centering
    \includegraphics[width=\textwidth]{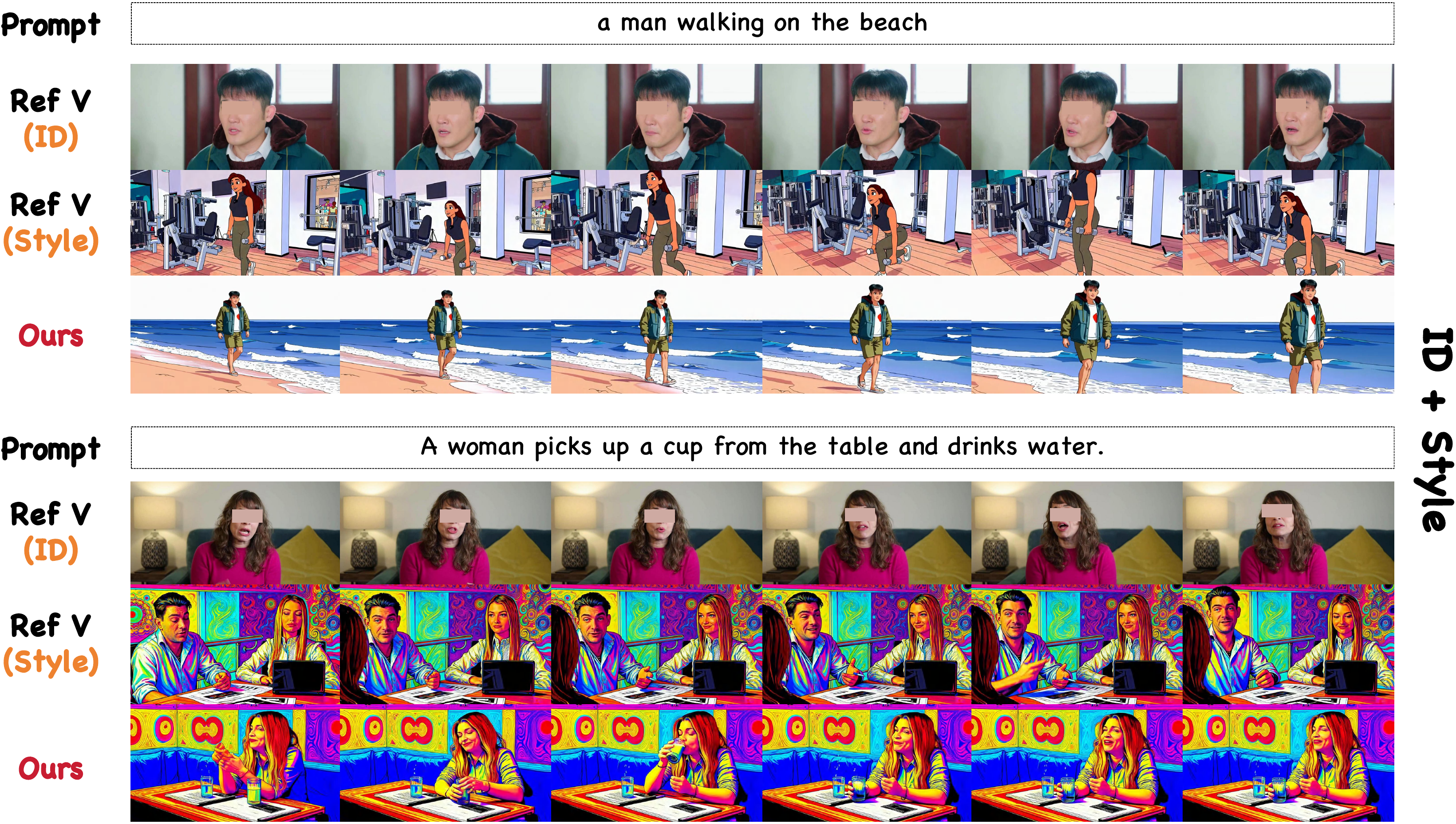}
    \caption{Video transfer combination results (Set 2).}
    \label{fig:comb-s2}
\end{figure*}

By concatenating, respectively, the reference video tokens and the MLLM tokens across different tasks, our approach enables seamless combination of multiple video transfer operations. Fig.~\ref{fig:comb-s1} and Fig.~\ref{fig:comb-s2} demonstrate the superiority of our method in handling entirely unseen task combinations, highlighting its strong generalization capability.

\end{document}